\documentclass[journal]{IEEEtran}
\newif\ifblind
\blindfalse      

\usepackage{bm}
\usepackage{booktabs}
\usepackage[table]{xcolor}
\definecolor{oursteal}{HTML}{2A9D8F}\colorlet{oursrow}{oursteal!22}
\definecolor{fcfspurple}{HTML}{E75480}\colorlet{failcell}{fcfspurple!12}
\definecolor{groupteal}{HTML}{4BA697}\colorlet{grouprow}{groupteal!8}

\usepackage{array}
\usepackage{makecell}
\usepackage{multirow}
\usepackage{algorithm}
\usepackage{algpseudocode}

\usepackage{amsmath, amssymb, amsthm, amsfonts}
\newtheorem{theorem}{Theorem}
\newtheorem{lemma}{Lemma}
\newtheorem{proposition}{Proposition}
\newtheorem{corollary}{Corollary}
\newtheorem{definition}{Definition}

\newtheorem{remark}{Remark}
\usepackage[flushleft]{threeparttable}
\usepackage{graphicx}
\usepackage{hyperref}
\usepackage{tabularx}
\usepackage{cite}
\graphicspath{{figs/}}

\usepackage{tikz}
\usetikzlibrary{arrows.meta}
\definecolor{orRed}{HTML}{E75480}\definecolor{roadexp}{HTML}{DADADA}
\definecolor{orTeal}{HTML}{2A9D8F}
\definecolor{orBlue}{HTML}{3B6FB6}
\definecolor{orPur}{RGB}{126,106,158}
\tikzset{orroad/.style={line width=5pt,color=roadexp,line cap=butt},
  orh/.style={fill=orTeal,draw=none,rounded corners=.6pt,minimum width=.46cm,minimum height=.22cm,inner sep=0pt},
  orv/.style={fill=orTeal,draw=none,rounded corners=.6pt,minimum width=.22cm,minimum height=.46cm,inner sep=0pt},
  ornew/.style={orh,fill=orBlue},
  orhole/.style={draw=orTeal,line width=.9pt,circle,minimum size=.52cm,inner sep=0pt},
  orarr/.style={-{Latex[length=4pt]},line width=.7pt,black!60},
  orchain/.style={-{Latex[length=4pt]},line width=.8pt,orRed},
  orbox/.style={draw=black!60,rounded corners=2pt,fill=black!4,inner sep=3pt,line width=.4pt,align=center,font=\scriptsize}}
\newcommand{\nx}{\operatorname{nx}}           

\title{OpenRing: Deadlock-Free Execution of Any Planner on AGV Floors with One-Way Lanes}

\ifblind
\author{}
\else
\author{Zhihao Lin, Anh-Tu~Nguyen,~\IEEEmembership{Senior Member,~IEEE,} Hui Zhang,~\IEEEmembership{Fellow,~IEEE,} and Yoshinobu Kawahara%
\thanks{This work was supported in part by the Japan Society for the Promotion of Science (JSPS)
through the JSPS Postdoctoral Fellowship for Research in Japan (Fellow ID P26385), in part by
JSPS KAKENHI Grant Number JP26H02498, and in part by the Japan Science and Technology Agency (JST)
Moonshot R\&D Program under Grant No. JPMJMS263K.
(Corresponding author: Yoshinobu Kawahara.)}
\thanks{Zhihao Lin is with the Graduate School of Information Science and Technology, The University of Osaka,
Suita, Osaka 565-0871, Japan (e-mail: zhihao.lin@ist.osaka-u.ac.jp).}%
\thanks{Anh-Tu Nguyen is with the LAMIH UMR CNRS 8201 Laboratory, Universit\'e Polytechnique Hauts-de-France, and INSA Hauts-de-France, Valenciennes 59313, France (e-mail: tnguyen@uphf.fr).}%
\thanks{Hui Zhang is with the School of Transportation Science and Engineering, Beihang University, Beijing 100191, China (e-mail: huizhang285@buaa.edu.cn).}
\thanks{Yoshinobu Kawahara is with the Graduate School of Information Science and Technology, The University of Osaka, Suita, Osaka 565-0871, Japan, and also with the RIKEN Center for Advanced Intelligence Project, Tokyo 103-0027,
Japan (e-mail: kawahara@ist.osaka-u.ac.jp).}
}
\fi

\usepackage{stfloats}   
\begin{document}
\maketitle

\begin{abstract}
Lifelong multi-agent path finding replans routes for automated guided vehicles, but deviations can invalidate a plan. With one-way lanes, a late vehicle or lost planner can close a ring of full segments. We separate how fast the floor runs from whether it keeps running. In OpenRing a planner proposes routes and an order, a local guard commits only moves and route changes that cannot close such a ring, and the order binds for a bounded time. Despite arbitrary proposals or planner loss, no ring closes. With two-place segments, traffic can continue under disturbances; progress is guaranteed once they end, and all vehicles exit if entries also stop. Among rules that reroute only on a proposal, every guard veto is necessary, and a clean plan keeps its crossing orders. The guarantee has a price: no rule that never stores a vehicle in a crossing under arbitrary stops carries more than $B/T_s$ per segment of $B$ places and traversal time $T_s$. With binding orders, three planners deadlock in 80 of 126 runs across eight floors. Inside OpenRing, none of the 126 runs deadlocks. OpenRing's closed-loop throughput rivals or exceeds the best baseline's. Physics simulations show no deadlocks or overlaps.
\end{abstract}


\section{Introduction}\label{sec:intro}

\IEEEPARstart{F}{leets} of automated guided vehicles (AGVs) are coordinated by plans. Lifelong multi-agent path finding (MAPF) replans
routes and an order of passage in a rolling window while vehicles keep arriving~\cite{li2021lifelong}, and an execution layer keeps the planned
order at every shared location~\cite{honig2019persistent,berndt2024receding,su2024bidirectional}. These methods reach high
throughput when the floor behaves as planned, and the working assumption is that a better plan makes a better floor.

\begin{figure}[!t]
\centering
\begin{minipage}[t]{0.49\columnwidth}\centering\includegraphics[width=\linewidth]{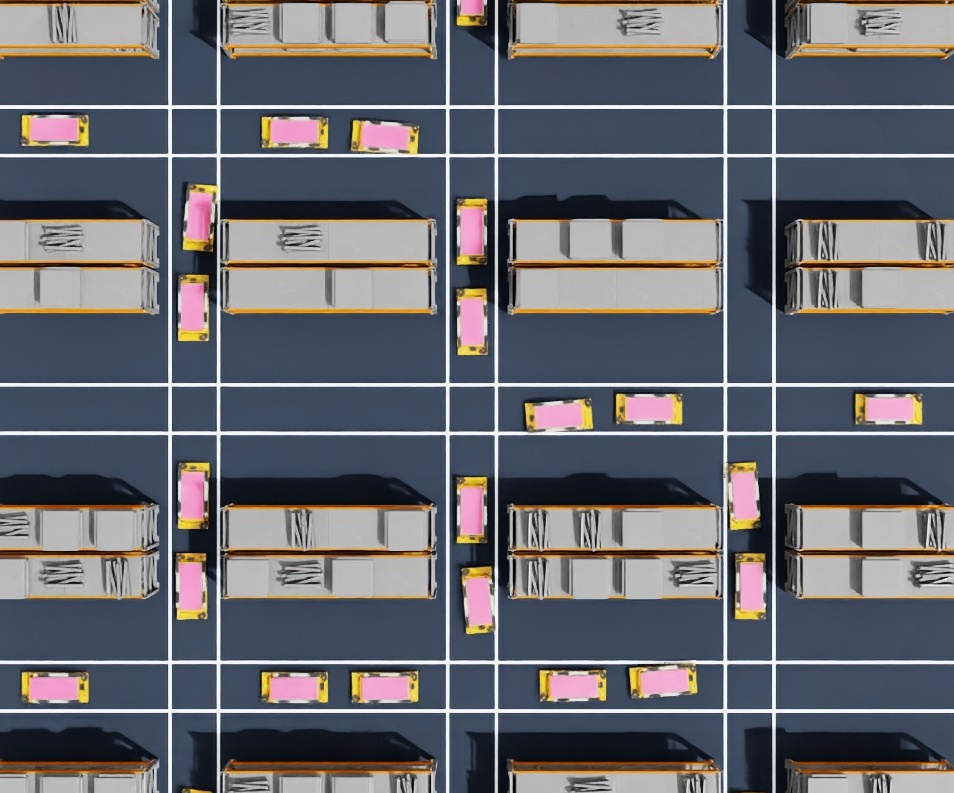}\\[2pt]{\footnotesize (a) First-come service}\end{minipage}\hfill
\begin{minipage}[t]{0.49\columnwidth}\centering\includegraphics[width=\linewidth]{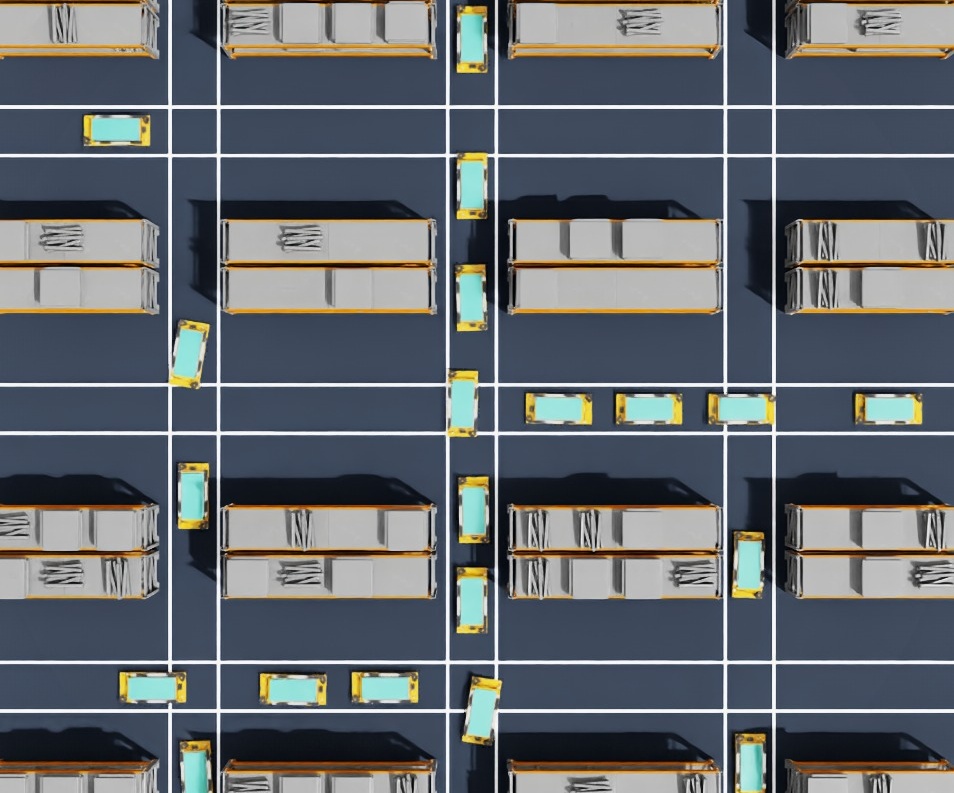}\\[2pt]{\footnotesize (b) OpenRing}\end{minipage}
\caption{The same arrivals in physics simulation. (a) Rings of full segments have closed, and the
pink AGVs never move. (b) AGVs keep moving.}
\label{fig:teaser}
\end{figure}

Yet a plan is valid only while the vehicles keep to it. Take the state that every floor must avoid
(Fig.~\ref{fig:teaser}(a)). Every aisle segment on a cycle is full, and every vehicle at the head of a segment waits for
space that only the vehicle ahead can free. Ordinary congestion drains once demand falls; a closed ring of fixed routes never does.
Warehouse floors invite it. Their aisles are narrow and usually one-way, the floor is a grid of crossings joined by short
segments, and with two places per segment eight AGVs in the wrong places close a block, however few vehicles the floor
holds in total. Real vehicles stop for picks, wait for people and report their state late, and a planner may be slow,
wrong or temporarily absent. Two things then go wrong. Replanning from a state the plan did not foresee can walk the floor
into a closed ring, because the planner looks a bounded time ahead and does not reason about storage. And keeping the
planned order strictly is no cure: an order that has become inconsistent makes vehicles wait for each other with room on
every segment, a deadlock that involves no full segment at all.

The difficulty lies in where safety is decided. When the plan decides it, every mistake of the plan is a risk to the
floor. The opposite tradition is safe by construction. Sequential resource allocation has shown that when every resource holds
at least two units, a one-step look-ahead avoids deadlock and is maximally
permissive~\cite{xing1996deadlock,reveliotis1997polynomial,lawley2001deadlock}. Interconnection networks keep a free buffer on each
ring for the same reason~\cite{puente2001adaptive}. These results concern the logic of allocation and say little about what it costs in throughput. Neither
tradition lets a planner decide almost everything while the guarantee survives its mistakes, and neither says what such a guarantee costs.

We therefore separate efficiency from safety. In OpenRing, a planner proposes routes and an order of passage. A local guard checks each proposed move or route change to keep a free place on every ring of waiting AGVs. The planner is free to propose, but cannot close a ring. This separation raises three questions.
\begin{itemize}
\item[(Q1)] Can safety hold under any planner, even one that changes routes while vehicles drive?
\item[(Q2)] How can a plan guide the floor without ever stopping it?
\item[(Q3)] What does a guarantee cost, and what must a method give up to do better?
\end{itemize}

We answer them on a model of FIFO segments with committed moves, and test the answers in a closed loop with acceleration
limits and full vehicle bodies, and in physics simulation of a commercial differential-drive AGV. OpenRing is not a new
planner but an interface: any planner plugs in unchanged and inherits the guarantees, with no proof of its own. It plays for execution the part LaCAM~\cite{okumura2023lacam} plays for search: LaCAM turns PIBT~\cite{okumura2022pibt}, which can fail, into a complete search, and OpenRing turns planners that deadlock when their order binds into a floor that never does. Executed alone with a binding order, PBS, PIBT and LaCAM deadlock in 80 of 126 runs on eight floors; inside OpenRing none of the 126 runs deadlocks. In most other experiments the planner is that of windowed replanning, so that the comparison isolates what the guard and the order expiry add. The contributions are as follows.

\begin{itemize}
\item \textbf{Safety under any planner (Q1).}
Route changes pass the same check as moves, so no ring of
full segments forms, whatever routes and orders are proposed.
Among rules that change routes only through proposals,
every veto of Look and $\Psi$ is one that rules surviving
planner loss must make, and no condition of OpenRing
can be dropped.

\item \textbf{Orders as advice (Q2).}
A binding order can stop the floor with room left, so the
order expires after a bounded time. The floor then keeps
moving, clears, and survives the loss of the planner,
while a clean plan keeps its order at every crossing.

\item \textbf{The price of a guarantee (Q3).}
Under arbitrary stops, no rule that keeps AGVs out of
crossings carries more than segment storage over its
traversal time, and the bound is tight on a lane of
the cell model without crossing delay.

\item \textbf{Planner interchangeability across eight floor layouts.}
PBS, PIBT and LaCAM plug into OpenRing unchanged. They deadlock in 80 of 126 runs alone and none inside OpenRing.
With stops and delayed observations, OpenRing delivers more AGVs than windowed replanning in all 12 paired runs.
We observe no deadlock or overlap in Isaac Sim or on a grid with up to 1108 AGVs at once. In Isaac Sim, traffic continues after planner shutdown.
\end{itemize}

Section~\ref{sec:related} reviews related work and
Section~\ref{sec:model} the model.
Sections~\ref{sec:rules} and~\ref{sec:exec}
describe the guard and execution;
Sections~\ref{sec:openring} and~\ref{sec:price}
establish the guarantees and their cost.
Section~\ref{sec:exp} presents experiments.
Proofs are in the Appendix.
\section{Related Work}\label{sec:related}

\begin{table}[t]
\centering\footnotesize
\begin{threeparttable}
\caption{OpenRing among the traditions it draws on.}
\label{tab:traditions}
\setlength{\tabcolsep}{3pt}
\begin{tabularx}{\columnwidth}{Xlll}
\toprule
Tradition & Guarantee & Any planner & Price\\
\midrule
AGV zones and windows & restricted states & no & --\\
Plan execution & planned order kept & no & --\\
Deadlock avoidance & no deadlock & no & --\\
Liveness supervisors & live, unit zones & own scheduler & --\\
\rowcolor{oursrow} OpenRing & safe, live, clearing & yes & $B/T_s$\\
\bottomrule
\end{tabularx}

\begin{tablenotes}\footnotesize
\item --: no throughput bound stated. Unit zones: each zone holds one vehicle.
\end{tablenotes}
\end{threeparttable}
\end{table}

\emph{AGV traffic management.} Industrial fleets reserve time windows on the guide-path
graph~\cite{smolicrocak2010time}, partition the floor into zones that hold one vehicle each and grant zones by
supervisory rules~\cite{draganjac2016decentralized,pratissoli2024hierarchical}, or plan routes centrally with priorities~\cite{bonetti2026traffic}, by decomposition~\cite{nishi2005distributed} or by distributed task allocation~\cite{camisa2023multi}. Deadlock freedom in these systems is usually obtained by restricting
the admitted states. Sequential resource allocation systems give the theory behind such restrictions: safety is
NP-hard in general, but becomes easy, with a one-step look-ahead as the maximally permissive policy, when every resource
has capacity at least two or a single predecessor or successor~\cite{xing1996deadlock,reveliotis1997polynomial,fanti1997event,lawley2001deadlock}.
Petri nets~\cite{wu2005modeling}, zone-controlled guide paths~\cite{roszkowska2008liveness,reveliotis2000conflict} and free-ranging vehicles~\cite{reveliotis2011conflict} follow the same line; for dynamically routed zones that hold one vehicle each, liveness was analysed, assessed and enforced in polynomial time~\cite{reveliotis2020state,reveliotis2020efficient,reveliotis2021polynomial} and kept inside model predictive schemes~\cite{daugherty2019optimized,reveliotis2020mpc}. Where a zone holds one vehicle, deciding liveness is a problem of its own; with storage for two it reduces to one walk. We build on the look-ahead principle from this line of work. Those supervisors sit below their own scheduler, on zones
that hold one vehicle. OpenRing instead sits below any
planner on segments that store several: it checks route
changes for deadlock, makes planned orders temporary,
keeps the floor clearable after planner loss, and
quantifies the throughput cost of the guarantee (Table~\ref{tab:traditions}).


\emph{Execution of multi-agent plans.} Lifelong multi-agent path finding plans warehouse traffic in discrete
time~\cite{li2021lifelong}, also with kinematic constraints~\cite{ma2019lifelong} and adapted to AGV motion~\cite{nagai2026gridworlds}. Its execution layer keeps the planned order at every shared location in an
action dependency graph~\cite{honig2016kinematic,honig2019persistent}; switchable dependencies re-order pairs of agents online by a
receding-horizon mixed-integer program~\cite{berndt2024receding}, and bidirectional temporal plan graphs allow either
passing order where the plan permits it~\cite{su2024bidirectional}. Plans robust to $k$ steps of delay~\cite{atzmon2020robust} or to delay probabilities~\cite{ma2017delay} absorb bounded lateness, but not arbitrary stops. Time-independent plans remove timing
altogether~\cite{okumura2021time,okumura2023otimapp}, offline plans can tolerate agents that crash~\cite{okumura2023fault}, and priority inheritance resolves local blocking without a plan~\cite{okumura2022pibt}.
Decentralized policies resolve conflicts safely~\cite{pallottino2007decentralized,zhou2017fast}, critical sections keep coordination with online goals safe and live~\cite{mannucci2021provably}, and kinodynamic and coverage planners deconflict whole trajectories or grid paths~\cite{tang2025stgcs,moldagalieva2026dbecbs,moldagalieva2026dblacam,zhang2026denoising,tang2025lsmcpp}. These methods trust a plan or path set; OpenRing treats a plan as advice, and we compare
with plan execution in a continuous closed loop.

\emph{Deadlock in interconnection networks.} Routing in multiprocessor networks is deadlock-free if the channel
dependency graph is acyclic~\cite{dally1987deadlock}, and bubble flow control keeps one free buffer
per ring~\cite{puente2001adaptive}. AGVs differ from packets in two ways that matter here: segments are FIFO without virtual
channels, and a passage occupies a crossing for a time that depends on whether the crossing vehicles share an aisle.

\emph{Safety filters.} A safety filter changes a nominal command only as far as a safety set requires, as control barrier functions do for continuous systems~\cite{ames2017cbf,wabersich2023filter} and for robot teams~\cite{wang2017safety}, where such filters can themselves deadlock the robots~\cite{zhang2025deadlock}; supervisory control gives the discrete counterpart~\cite{ramadge1987supervisory}. Algorithms with predictions trade trust in advice against worst-case loss~\cite{lykouris2021competitive}. OpenRing brings these ideas to a fleet, with a planner as the nominal command.


\section{Problem Formulation}\label{sec:model}

\subsection{Floor and vehicles}

\begin{figure}[t]
\centering
\begin{tikzpicture}[x=1cm,y=1cm,font=\footnotesize,
  road/.style={line width=5pt,color=roadexp,line cap=butt},
  h/.style={fill=orTeal,draw=none,rounded corners=.6pt,minimum width=.46cm,minimum height=.22cm,inner sep=0pt},
  v/.style={fill=orTeal,draw=none,rounded corners=.6pt,minimum width=.22cm,minimum height=.46cm,inner sep=0pt},
  nx/.style={-{Latex[length=4pt]},line width=.7pt,orRed}]
\begin{scope}
\draw[road] (-.7,0) -- (3.2,0); \draw[road] (0,-.7) -- (0,2.3); \draw[road] (2.5,-.7) -- (2.5,2.3); \draw[road] (-.7,1.6) -- (3.2,1.6);
\draw[black!60,line width=.5pt] (.2,1.6) ++(0,0);
\node[h] at (1.05,0) {}; \node[h] at (1.75,0) {};
\draw[black!55,line width=.5pt,dashed] (2.17,-.16) -- (2.17,.16);
\node[below] at (1.25,-.14) {segment $s$, $B=2$};
\node[circle,draw=black!55,line width=.4pt,minimum size=.55cm,inner sep=0pt] at (2.5,0) {};
\node[right] at (2.78,-.34) {crossing};
\node[above] at (1.6,.12) {head};
\draw[-{Latex[length=3.5pt]},black!55,line width=.5pt] (1.98,.3) -- ++(.3,0);
\node[right] at (2.62,.5) {$\nx(s)$};
\draw[nx] (2.5,.08) -- (2.5,1.0);
\node at (1.25,-1.0) {(a) Segment, head, next segment};
\end{scope}
\begin{scope}[xshift=4.5cm]
\draw[road] (-.7,0) -- (2.9,0); \draw[road] (-.7,1.6) -- (2.9,1.6); \draw[road] (0,-.7) -- (0,2.3); \draw[road] (2.2,-.7) -- (2.2,2.3);
\node[h,fill=orRed] at (.7,0) {}; \node[h,fill=orRed] at (1.5,0) {}; \node[h,fill=orRed] at (.7,1.6) {}; \node[h,fill=orRed] at (1.5,1.6) {};
\node[v,fill=orRed] at (2.2,.5) {}; \node[v,fill=orRed] at (2.2,1.15) {}; \node[v,fill=orRed] at (0,.5) {}; \node[v,fill=orRed] at (0,1.15) {};
\draw[nx] (1.75,.14) -- (2.05,.14);
\draw[nx] (2.06,1.35) -- (2.06,1.5);
\draw[nx] (.45,1.46) -- (.15,1.46);
\draw[nx] (.14,.3) -- (.14,.14);
\node[align=center,text=orRed] at (1.1,.8) {every head\\waits};
\node at (1.1,-1.0) {(b) Full wait cycle};
\end{scope}
\end{tikzpicture}%
\vspace{-3mm}\caption{Model. (a) A segment stores $B$ AGVs and its head wants the next segment. (b) A full wait cycle.}
\label{fig:model}
\end{figure}
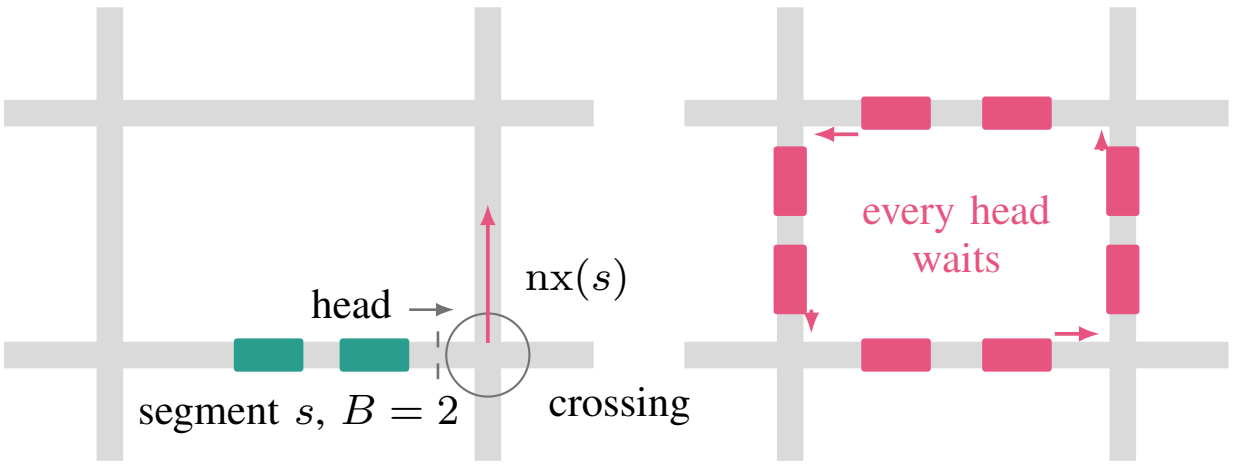

The floor is a directed graph
\begin{equation}\label{eq:floor}
\mathcal{G}=(\mathcal{C},\mathcal{S}),\qquad \mathcal{S}=\mathcal{S}_{\mathrm{in}}\cup\mathcal{S}_{\mathrm{int}}\cup\mathcal{S}_{\mathrm{out}},
\end{equation}
whose nodes $c\in\mathcal{C}$ are crossings and whose edges are one-way aisle segments: entry, internal and exit segments. A two-way aisle with a lane per direction is a pair of opposite segments, and roundabouts and loops are chains of segments, so the model is not limited to one-way grids.
Every internal segment $s$ runs from its feeding crossing $\mathrm{up}(s)$ to $\mathrm{dn}(s)$, is FIFO and stores at most
$B\ge2$ stopped AGVs (Fig.~\ref{fig:model}a). An AGV $i\in\mathcal{I}$ follows a route, a finite walk in $\mathcal{G}$ from an entry to an exit,
\begin{equation}\label{eq:route}
\rho_i=(s_i^1,\dots,s_i^{L_i}),\quad \mathrm{dn}(s_i^{m})=\mathrm{up}(s_i^{m+1}),\quad L_i\le L.
\end{equation}
The bound $L$ also holds for the remaining part of every route proposed later.

Each AGV is a rectangle of length $\ell$ and width $w$ with state
\begin{equation}\label{eq:state}
q_i(t)=[x_i,y_i,\theta_i,v_i]^\top,\qquad 0\le v_i\le V,\quad |\dot v_i|\le a_{\max},
\end{equation}
and footprint $\mathcal{B}(q_i)\subset\mathbb{R}^2$. Bodies must stay disjoint,
\begin{equation}\label{eq:body}
\mathcal{B}(q_i(t))\cap\mathcal{B}(q_j(t))=\emptyset,\qquad \forall\, i\ne j,\ \forall\, t.
\end{equation}
A passage blocks its crossing for a time $\Delta$, and two AGVs
that share a segment at the crossing pass at least $h_0\ge\Delta$ apart. An AGV is never held inside a crossing; an AGV that turns there rotates in place as part of its passage.

\subsection{Discrete state and wait graph}

Decisions are made on a discrete state. A \emph{commitment} $(i,s\to u)$ irrevocably allows AGV $i$ to pass from $s$ into
$u$, and counts $i$ in $u$ from that instant. Let $\sigma_i(t)\in\mathcal{S}$ be the segment in which $i$ is counted, and
\begin{equation}\label{eq:occ}
n_s=\bigl|\{i:\sigma_i=s\}\bigr|,\qquad \mathcal{F}=\{s\in\mathcal{S}_{\mathrm{int}}:n_s=B\}
\end{equation}
the occupancy and the set of full segments. Let $h(s)$ be the head of a non-empty segment, the first AGV counted in it in FIFO
order, and $\nx(s)$ the next segment on the route of $h(s)$. The discrete state is
\begin{equation}\label{eq:zstate}
z(t)=\bigl(\sigma(t),\rho(t)\bigr)\in\mathcal{Z}.
\end{equation}
The \emph{wait graph} of $z$ is the functional graph of full segments,
\begin{equation}\label{eq:wait}
\mathcal{W}(z)=\bigl\{(s,\nx(s)):s\in\mathcal{F},\ \nx(s)\in\mathcal{F}\bigr\}.
\end{equation}

\begin{definition}[Full wait cycle]\label{def:fwc}
A \emph{full wait cycle} is a directed cycle $u_0\to u_1\to\dots\to u_k=u_0$ of $\mathcal{W}(z)$. The safe set is
\begin{equation}\label{eq:safe}
\mathcal{Z}_{\mathrm{safe}}=\{z\in\mathcal{Z}:\ \mathcal{W}(z)\ \text{is acyclic}\}.
\end{equation}
\end{definition}

No AGV on a full wait cycle (Fig.~\ref{fig:model}b) can move while routes stay fixed: each head waits for a full segment whose own head waits in turn.
With routes fixed and moves made one AGV at a time, Lemma~\ref{lem:stuck} shows that every deadlock of the storage contains one; a binding order can stop the floor without one (Proposition~\ref{prop:hard}(b)). The state evolves by events,
\begin{equation}\label{eq:event}
z^{+}=\Gamma(z,e),\qquad e\in\{\mathsf{commit},\mathsf{enter},\mathsf{exit},\mathsf{reroute}\}.
\end{equation}
An exit is the commitment of an AGV into its exit segment; the AGV then leaves without further events and is no longer counted.

\subsection{Demand, planner and disturbances}

AGVs arrive at the entries as Poisson streams, and $\lambda_s$ denotes the flow on segment $s$. Arrival rates are stated as a load, a fraction of the
per-crossing bound
\begin{equation}
\eta_c=(\lambda_a+\lambda_b)\Delta+(h_0-\Delta)\,|\lambda_a-\lambda_b|,
\label{eq:bound}
\end{equation}
where $\lambda_a,\lambda_b$ are the flows on the two incoming segments of $c$; no rule exceeds $\eta_c=1$ (Lemma~\ref{lem:bound}). Every crossing of our floors has at most two incoming segments.
The demand joins entries to exits by routes that may turn several times. A \emph{planner} is any map
\begin{equation}\label{eq:planner}
\Pi:\ z(t)\ \mapsto\ \bigl(\hat\rho(t),\ \{\hat\prec_c(t)\}_{c\in\mathcal{C}}\bigr),
\end{equation}
which proposes routes $\hat\rho$ and an order of passage $\hat\prec_c$ at each crossing. Nothing is assumed about $\Pi$: its
proposals may be wrong, late, inconsistent or absent, and it may use time and the whole history of the floor. A proposed route continues from the committed target of its AGV, ends at the exit of the AGV and has at most $L$ segments. Let $\mathcal{P}$ denote all planners and $\mathcal{D}$ all
disturbances, which stop uncommitted AGVs for arbitrary finite times or delay the state seen by crossings.

\subsection{Problem statement}\label{sec:problem}

A \emph{rule} is a map $\mu$ that decides, from $z$ and the proposals, which events are applied. Its throughput is
\begin{equation}\label{eq:thr}
\Lambda(\mu;\Pi,d)=\liminf_{T\to\infty}\frac{1}{T}\,\bigl|\{i\in\mathcal{I}:\ i\ \text{exits by}\ T\}\bigr|.
\end{equation}
We seek
\begin{subequations}\label{eq:problem}
\begin{align}
\max_{\mu}\quad
& \Lambda(\mu;\Pi,d) \label{eq:p0}\\
\mathrm{s.t.}\quad
& z(t)\in\mathcal Z_{\mathrm{safe}},
  \ \eqref{eq:body}, \label{eq:p1}\\
& \text{progress within }\bar T, \label{eq:p2}\\
& \mathcal I(T_1)=\emptyset,\quad
  T_1<\infty. \label{eq:p3}
\end{align}
\end{subequations}
where $\mathcal I(t)$ is the set of AGVs on the floor.
Safety~\eqref{eq:p1} holds at all times;
progress~\eqref{eq:p2} requires a commitment or exit
within $\bar T$ after disturbances end whenever
$\mathcal I(t)\ne\emptyset$; and clearing~\eqref{eq:p3}
requires some finite $T_1$ if inflow and disturbances
end by $T_0$. All constraints hold for every
$(\Pi,d)\in\mathcal P\times\mathcal D$; the objective
is evaluated on the planner at hand, so a rule should
preserve good plans and degrade gracefully otherwise.
We also ask for the converse: the largest flow that any rule feasible for a class of disturbances can carry.
Sections~\ref{sec:rules}--\ref{sec:openring} construct a feasible $\mu$ (Corollary~\ref{thm:feas}), Section~\ref{sec:price} bounds $\Lambda$, and
Section~\ref{sec:exp} evaluates it.

\section{The Local Guard}\label{sec:rules}

Each crossing decides from its own heads and the wait chain that starts at each target. The guard is local in what it reads, not in where it runs: a
decision reads one chain of full segments, not the plan and not the rest of the floor, and the coordinator of
Section~\ref{sec:openring} evaluates it on the ledger. Its candidates are the heads of its incoming segments, and the request $r(i)$ of a candidate
$i$ is the earliest time it can reach the crossing. A \emph{move} $s\to s'$ commits the head of $s$ into $s'=\nx(s)$. It changes
only $n_s$, $n_{s'}$, the head of $s$, and the head of $s'$ if $s'$ was empty. First-come service admits moves into
a segment with room, and it can leave $\mathcal{Z}_{\mathrm{safe}}$ as soon as AGVs turn.

\subsection{One-step look-ahead}

Let $z'$ be the state after the move $s\to s'$. The guard follows the heads from the target,
\begin{equation}\label{eq:walk}
x_0=s',\qquad x_{k+1}=\nx'(x_k),
\end{equation}
where $\nx'$ is $\nx$ evaluated in $z'$, and it admits the move unless the chain closes through full segments:
\begin{equation}\label{eq:look}
\mathrm{Look}(z,s\to s')=
\begin{cases}
0, & n_{s'}=B\ \text{ or }\ \chi(z')=1,\\
1, & \text{otherwise,}
\end{cases}
\end{equation}
with $\chi(z')=1$ if the chain closes, that is, $x_k=s'$ for some $k\ge1$ with $x_0,\dots,x_{k-1}\in\mathcal{F}(z')$.
Because $\nx'$ is a function, \eqref{eq:walk} is a single chain, and Algorithm~\ref{alg:look} evaluates~\eqref{eq:look} by
following it. The walk stops at a segment with room, at an exit, or at $s$, which the move frees.

\begin{algorithm}[t]
\caption{Look-ahead $\mathrm{Look}(z,s\to s')$}\label{alg:look}
\begin{algorithmic}[1]
\Require state $z\in\mathcal{Z}_{\mathrm{safe}}$; move $s\to s'$
\If{$n_{s'}=B$} \Return 0 \EndIf
\If{$n_{s'}+1<B$} \Return 1 \EndIf
\State $x\gets s'$
\Loop
  \State $y\gets\nx'(x)$
  \If{$y=s'$} \Return 0 \EndIf
  \If{$y$ is an exit, $y=s$, or $n_y<B$} \Return 1 \EndIf
  \State $x\gets y$
\EndLoop
\end{algorithmic}
\end{algorithm}

The look-ahead is classical for resources of capacity at least two~\cite{xing1996deadlock,reveliotis1997polynomial,fanti1997event,lawley2001deadlock}.
New here are its FIFO form, in which only the head can leave a segment, where the classical models let any process leave a buffer, and for $B\ge2$ a move into a nonempty segment leaves the head unchanged; the
route check, the expiry of the order, the necessity of the guard after the planner is lost
(Theorem~\ref{thm:least}) and its price (Theorem~\ref{thm:flow}). In FIFO form:

\begin{proposition}[Local guard]\label{thm:look}
Let $B\ge2$ and let routes be fixed. If moves are made one at a time, each with $\mathrm{Look}=1$ on the current state, and each crossing serves them in any work-conserving order, one that never leaves the crossing idle while it has such a move,
then (i) $\mathcal{Z}_{\mathrm{safe}}$ is invariant, and (ii) whenever AGVs are on the floor, a move or an exit
occurs in finite time.
\end{proposition}

The proof is in Appendix~\ref{app:guard}. It rests on two facts
about a single state. If no full wait cycle exists, some head can move (Lemma~\ref{lem:stuck}). And if a move into a
segment with room is refused, another incoming segment of the same crossing holds a head whose move is admitted
(Lemma~\ref{lem:alt}). The latter is where $B\ge2$ enters. Neither fact uses the grid, so Proposition~\ref{thm:look} holds on any network of FIFO
segments.

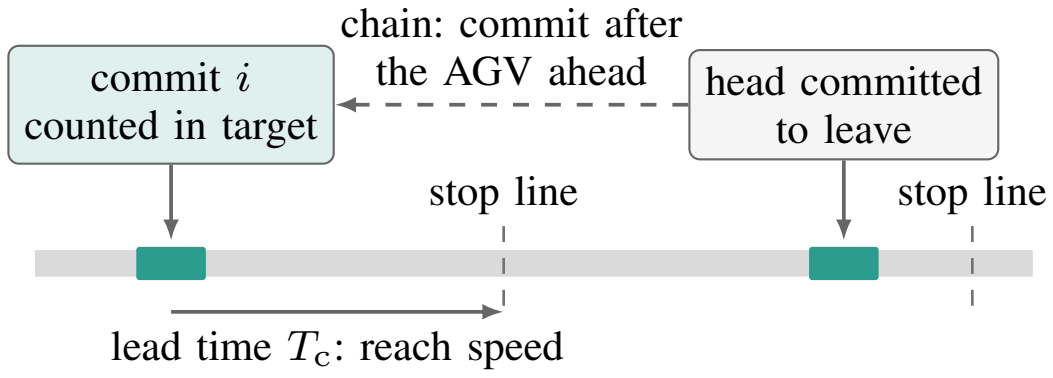
\begin{figure}[t]
\centering
\begin{tikzpicture}[x=1cm,y=1cm,font=\footnotesize,every node/.append style={font=\footnotesize}]
\draw[orroad] (0,0) -- (6.6,0);
\draw[black!55,dashed,line width=.5pt] (3.1,-.28) -- (3.1,.28); \node at (3.1,.45) {stop line};
\draw[black!55,dashed,line width=.5pt] (6.2,-.28) -- (6.2,.28); \node at (6.2,.45) {stop line};
\node[orh] at (.9,0) {}; \node[orh] at (5.35,0) {};
\node[orbox,font=\footnotesize,fill=orTeal!15] (c) at (.9,1.05) {commit $i$\\counted in target};
\draw[orarr] (c) -- (.9,.15);
\draw[orarr] (.9,-.32) -- node[below,text=black] {lead time $T_{\mathrm{c}}$: reach speed} (3.1,-.32);
\node[orbox,font=\footnotesize] (d) at (5.35,1.05) {head committed\\to leave};
\draw[orarr] (d) -- (5.35,.15);
\draw[orarr,dashed] (d.west) -- node[above,text=black,align=center] {chain: commit after\\the AGV ahead} (c.east);
\end{tikzpicture}%
\vspace{-3mm}\caption{Commitment. A move is committed a lead time ahead, so the AGV crosses at speed; a commitment may rest on a head that is committed to leave.}
\label{fig:commit}
\end{figure}

\section{Execution}\label{sec:exec}

A rule that decides at the instant of passage forces every waiting AGV to start from standstill at the stop line. We
therefore separate the commitment of~\eqref{eq:occ} from the passage.

\emph{Commitment.} A crossing commits a passage a lead time $T_{\mathrm{c}}$ before it happens, so that an AGV can cross at speed. A committed passage is kept. A crossing serves its committed passages in the order of commitment, and an AGV holds at most one commitment that it has not yet executed.

\emph{Chained commitment.} A committed AGV is counted in its target at once, so a segment can have counted room while it is still physically full (Fig.~\ref{fig:commit}). The crossing may then commit another AGV for later passage, propagating commitments upstream through the queue. A crossing uses the state of its own incoming segments, the commitments of the crossings
directly downstream, and the walk of Algorithm~\ref{alg:look}.

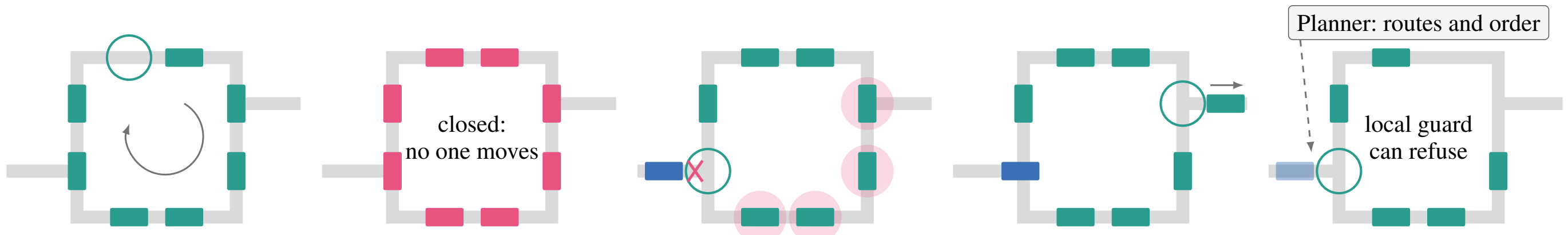
\begin{figure*}[t]
\centering
\resizebox{\textwidth}{!}{%
\begin{tikzpicture}[x=1cm,y=1cm,font=\fontsize{11.2}{12.6}\selectfont,
  road/.style={line width=6pt,color=roadexp,line cap=butt},
  h/.style={fill=orTeal,draw=none,rounded corners=.8pt,minimum width=.62cm,minimum height=.30cm,inner sep=0pt},
  v/.style={fill=orTeal,draw=none,rounded corners=.8pt,minimum width=.30cm,minimum height=.62cm,inner sep=0pt},
  p/.style={fill=orBlue,draw=none,rounded corners=.8pt,minimum width=.62cm,minimum height=.30cm,inner sep=0pt},
  hole/.style={draw=orTeal,line width=1.1pt,circle,minimum size=.72cm,inner sep=0pt},
  halo/.style={fill=orRed,fill opacity=.22,draw=none,circle,minimum size=.86cm,inner sep=0pt},
  note/.style={align=center,font=\fontsize{11.2}{12.6}\selectfont},
  subcap/.style={align=center,font=\small},
  arr/.style={-{Latex[length=5pt]},line width=.8pt}]
\newcommand{\ringbase}{%
  \draw[road] (0,0) rectangle (2.6,2.6);
  \draw[road] (-1.15,.75) -- (0,.75);
  \draw[road] (2.6,1.85) -- (3.65,1.85);}
\begin{scope}
\ringbase
\node[hole] at (.85,2.6) {};
\node[h] at (1.75,2.6) {}; \node[v] at (2.6,1.85) {}; \node[v] at (2.6,.75) {}; \node[h] at (1.75,0) {}; \node[h] at (.85,0) {}; \node[v] at (0,.75) {}; \node[v] at (0,1.85) {};
\draw[arr,black!55] (1.75,1.85) arc (60:-200:.62);
\node[subcap] at (1.3,-.75) {(a) One free place:\\every vehicle can still move};
\end{scope}
\begin{scope}[xshift=5.15cm]
\ringbase
\node[h,fill=orRed] at (.85,2.6) {}; \node[h,fill=orRed] at (1.75,2.6) {}; \node[v,fill=orRed] at (2.6,1.85) {}; \node[v,fill=orRed] at (2.6,.75) {}; \node[h,fill=orRed] at (1.75,0) {}; \node[h,fill=orRed] at (.85,0) {}; \node[v,fill=orRed] at (0,.75) {}; \node[v,fill=orRed] at (0,1.85) {};
\node[note] at (1.3,1.3) {closed:\\no one moves};
\node[subcap] at (1.3,-.75) {(b) Without a guard the last place\\is taken: the floor stops for good};
\end{scope}
\begin{scope}[xshift=10.3cm]
\ringbase
\node[halo] at (2.6,1.85) {}; \node[halo] at (2.6,.75) {}; \node[halo] at (1.75,0) {}; \node[halo] at (.85,0) {};
\node[hole] at (0,.75) {};
\node[h] at (.85,2.6) {}; \node[h] at (1.75,2.6) {}; \node[v] at (2.6,1.85) {}; \node[v] at (2.6,.75) {}; \node[h] at (1.75,0) {}; \node[h] at (.85,0) {}; \node[v] at (0,1.85) {};
\node[p] at (-.72,.75) {};
\draw[orRed,line width=1.3pt] (-.33,.58) -- (-.09,.92) (-.33,.92) -- (-.09,.58);
\node[subcap] at (1.3,-.75) {(c) With the guard: the full chain\\would close, the move is refused};
\end{scope}
\begin{scope}[xshift=15.45cm]
\ringbase
\node[hole] at (2.6,1.85) {};
\node[h] at (.85,2.6) {}; \node[h] at (1.75,2.6) {}; \node[v] at (2.6,.75) {}; \node[h] at (1.75,0) {}; \node[h] at (.85,0) {}; \node[v] at (0,1.85) {};
\node[h] at (3.3,1.85) {};
\node[p,rotate=0] at (-.05,.75) {};
\draw[arr,black!55] (3.05,2.15) -- (3.6,2.15);
\node[subcap] at (1.3,-.75) {(d) A vehicle leaves:\\the same move is now admitted};
\end{scope}
\begin{scope}[xshift=20.6cm]
\ringbase
\node[draw=black!65,fill=black!4,rounded corners=2pt,inner sep=4pt,minimum width=2.7cm] (pl) at (1.3,3.17) {Planner: routes and order};
\node[hole] (g) at (0,.75) {};
\node[note] at (1.3,1.3) {local guard\\can refuse};
\node[h] at (.85,2.6) {}; \node[v] at (2.6,.75) {}; \node[h] at (1.75,0) {}; \node[h] at (.85,0) {}; \node[v] at (0,1.85) {};
\node[p,opacity=.45] at (-.72,.75) {};
\draw[arr,black!55,dashed] (pl.south west) ++(.2,0) -- (-.45,1.05);
\node[subcap] at (1.3,-.75) {(e) OpenRing: the plan proposes,\\the guard decides};
\end{scope}
\end{tikzpicture}}%
\vspace{-3mm}\caption{OpenRing in five steps. A ring keeps moving while it has one free place, and stops for good when the last one is taken,
so the guard refuses only that move. A wrong plan costs throughput, not safety.}
\label{fig:idea}
\end{figure*}

\emph{The ledger and the execution contract.} The coordinator keeps the counted state~\eqref{eq:occ} as a ledger. A
commitment moves its AGV from the source count to the target count at once, events are applied one at a time, and Look and
$\Psi$ read only the ledger. A delayed view of the floor therefore changes when a head is ready and when physical room is
confirmed, never the counts that Look reads. The vehicles execute the ledger under five conditions; (C1)--(C4) give safety and (C5) gives progress. Let $\gamma_i(t)$ be the free distance from the front of $i$ to the rear of the AGV ahead of it on its lane, $\gamma^\star(v)=\gamma_0+t_\gamma v$ the car-following gap at speed $v$, $\varepsilon_s(t)$ the free length from the entry of $s$ to the rear of the last AGV on $s$, $I_i^c=[a_i^c,b_i^c]$ the passage of $i$ through $c$, from its stop line until its body has left $c$, and $\mathcal{D}_c$ the disk swept by a rotation in place at $c$.
\begin{itemize}
\item[(C1)] A passage of $i$ from $s$ into $s'$ at $c$ satisfies $\varepsilon_{s'}(a_i^c)\ge\ell+\gamma_0+\epsilon$ for a slack $\epsilon>0$.
\item[(C2)] $\gamma_i(t)\ge\gamma^\star(v_i(t))$ for every AGV $i$ and time $t$, on the true positions.
\item[(C3)] Passages of $i\ne j$ at $c$ whose movements share a segment satisfy $|a_i^c-a_j^c|\ge h_0$.
\item[(C4)] Passages of $i\ne j$ at $c$ with different movements satisfy $I_i^c\cap I_j^c=\emptyset$, and every stop line of $c$ is placed so that a body waiting at it does not meet $\mathcal{D}_c$.
\item[(C5)] Let $r_i$ be the remaining distance along the
lane to the stop line. In the absence of disturbances,
while $r_i>0$, AGV $i$ has $\dot v_i=a_{\max}$ whenever
$
v_i<\min\!\left\{V,\frac{\gamma_i-\gamma_0}{t_\gamma},
                     \sqrt{2a_{\max}r_i}\right\},
$
with the gap term omitted if no AGV is ahead on the
segment. At the stop line, a committed AGV starts its
passage as soon as its turn comes and (C1)--(C4)
permit entry.
\end{itemize}
Under the segment-length condition of
Proposition~\ref{prop:d0}, (C5) yields the finite bound
$D_0$ used in Theorem~\ref{thm:or-live}.

\begin{lemma}[Committed occupancy]\label{lem:commit}
Let the coordinator keep the ledger, let the vehicles satisfy (C1)--(C4), and let disturbances stop only AGVs that have not passed their stop line.
Then Proposition~\ref{thm:look}(i) holds for the ledger, bodies stay disjoint~\eqref{eq:body}, and no AGV is held inside a
crossing.
\end{lemma}

An AGV inside a crossing belongs to neither segment physically. If the counts followed the physical entry instead, a
walk would see room in a segment that a committed AGV is about to fill and could admit the move that closes a cycle.

\section{OpenRing: Planning Above a Local Guard}\label{sec:openring}

The guard sees the chain ahead of one move, and a planner sees the whole floor. OpenRing gives each its task: the planner proposes, and the local rule decides (Fig.~\ref{fig:idea}).

\subsection{The rule}

All quantities refer to the counted state~\eqref{eq:zstate}. Events are applied one at a time by the coordinator of the floor,
and a commitment is never withdrawn.

For a crossing $c$, the admitted candidates are
\begin{equation}\label{eq:admit}
\begin{aligned}\mathcal{A}_c(z)=\bigl\{h(s):\ &\mathrm{dn}(s)=c,\ h(s)\ \text{ready},\\ &\mathrm{Look}(z,s\to\nx(s))=1\bigr\},\end{aligned}
\end{equation}
with $\mathrm{Look}$ from~\eqref{eq:look}; a head is ready when it holds no unexecuted commitment. A proposed route
$\hat\rho_i$ from~\eqref{eq:planner} is checked by
\begin{equation}\label{eq:reroute}
\Psi(z,i,\hat\rho_i)=
\begin{cases}
0, & i=h(x),\ x\in\mathcal{F}(z),\ z[\hat\rho_i]\notin\mathcal{Z}_{\mathrm{safe}},\\
1, & \text{otherwise,}
\end{cases}
\end{equation}
where $z[\hat\rho_i]$ is $z$ with the uncommitted part of $\rho_i$ replaced. Let $\kappa_i$ count the route changes of $i$, a proposal equal to the current route being no change, $\hat r_c(i)$ be
the rank of $i$ in $\hat\prec_c$, and $\tau_c(t)$ the time since the last commitment at $c$. OpenRing has two parameters: the limit $K$ on
route changes and the time $T_f$ for which an order may bind.

\begin{algorithm}[t]
\caption{OpenRing at crossing $c$}\label{alg:openring}
\begin{algorithmic}[1]
\Require state $z$; proposals $\hat\rho,\hat\prec_c$ (possibly none)
\If{a commitment of $c$ is pending} serve it at its time; \Return \EndIf
\For{each AGV $i$ with a new proposal $\hat\rho_i\ne\rho_i$ and $\kappa_i<K$, once per step}
  \If{$\Psi(z,i,\hat\rho_i)=1$} replace the uncommitted part of $\rho_i$; $\kappa_i\gets\kappa_i+1$ \EndIf
\EndFor
\State compute $\mathcal{A}_c(z)$ by~\eqref{eq:admit}
\State select $i_c^\star$ by~\eqref{eq:select}
\If{$i_c^\star$ exists} commit it; $\tau_c\gets0$ \EndIf
\end{algorithmic}
\end{algorithm}

\begin{definition}[OpenRing]\label{def:openring}
OpenRing is the rule $\mu^\star$ with the following events.
\begin{enumerate}
\item \emph{Commitment.} Only moves of $\mathcal{A}_c(z)$ are committed.
\item \emph{Route change.} $\rho_i$ is replaced if and only if $\Psi(z,i,\hat\rho_i)=1$ and $\kappa_i<K$.
\item \emph{Selection.} Crossing $c$ selects
\begin{equation}\label{eq:select}
i_c^\star=
\begin{cases}
\hat h_c\ \text{if}\ \hat h_c\in\mathcal{A}_c,\ \text{none otherwise}, & \tau_c<T_f,\\
\arg\min_{i\in\mathcal{A}_c}\ r(i), & \text{otherwise,}
\end{cases}
\end{equation}
where $\hat h_c$ is the AGV of the first passage of $\hat\prec_c$ not yet committed at $c$ and not refused by $\mathrm{Look}$. In the first case $c$ waits while $\hat h_c$ is not yet a candidate. When $\hat\prec_c$ is void, as after a loss of the planner, $c$ uses the second case at once.
\end{enumerate}
\end{definition}

Selection~\eqref{eq:select} never enlarges $\mathcal{A}_c$: the plan only chooses among legal moves. The time $T_f$
makes the order advice rather than a constraint. Algorithm~\ref{alg:openring} summarizes one step. Without a planner, OpenRing is first-come service with fixed routes.

\subsection{Safety}

We now check the constraints of~\eqref{eq:problem}. Safety is an invariance statement for the event map~\eqref{eq:event}.

\begin{theorem}[Invariance]\label{thm:or-safe}
For every event $e$ of~\eqref{eq:event} under Definition~\ref{def:openring} and every planner $\Pi\in\mathcal{P}$,
\begin{equation}\label{eq:inv}
z\in\mathcal{Z}_{\mathrm{safe}}\ \Longrightarrow\ \Gamma(z,e)\in\mathcal{Z}_{\mathrm{safe}}.
\end{equation}
\end{theorem}

\begin{figure}[t]
\centering
\begin{tikzpicture}[x=1cm,y=1cm,font=\footnotesize]
\begin{scope}[yshift=-.55cm]
\draw[orroad] (0,0) rectangle (2.0,1.3); \draw[orroad] (2.0,1.3) -- (3.0,1.3);
\foreach \x in {.55,1.45} {\node[orh] at (\x,0) {}; \node[orh] at (\x,1.3) {};}
\node[orv] at (0,.65) {}; \node[orv] at (2.0,.65) {};
\draw[orarr,dashed] (1.75,1.5) -- (2.8,1.5); \node[anchor=west] at (2.05,1.72) {keep route};
\draw[orchain] (1.8,1.1) -- (1.8,.85);
\draw[orRed,line width=1.1pt] (1.7,1.05) -- (1.9,.85) (1.7,.85) -- (1.9,1.05);
\end{scope}
\node[align=center,anchor=north,font=\scriptsize] at (1.3,-.82) {(a) A change that would close\\a ring is refused};
\begin{scope}[xshift=4.6cm]
\draw[orroad] (0,0) -- (3.2,0); \draw[orroad] (.8,-.55) -- (.8,1.1); \draw[orroad] (2.4,-.55) -- (2.4,1.1);
\node[orh] (X) at (1.9,0) {}; \node at (1.9,.26) {$X$};
\node[orv,fill=orBlue] (Y) at (.8,-.4) {}; \node at (.54,-.4) {$Y$};
\node[orhole,draw=black!50] at (.8,0) {}; \node[orhole,draw=black!50] at (2.4,0) {};
\node at (.55,.95) {$c_1$}; \node at (2.65,.95) {$c_2$};
\node[align=center] at (1.6,.78) {each waits\\for the other};
\node[align=center,anchor=north,font=\scriptsize] at (1.6,-.82) {(b) A binding order\\stops the floor};
\end{scope}
\end{tikzpicture}%
\vspace{-3mm}\caption{Why routes are checked and orders expire. (a) Route changes pass the same check as moves. (b) With room on every segment, an inconsistent order still stops the floor; after $T_f$ the order lapses.}
\label{fig:route}
\end{figure}
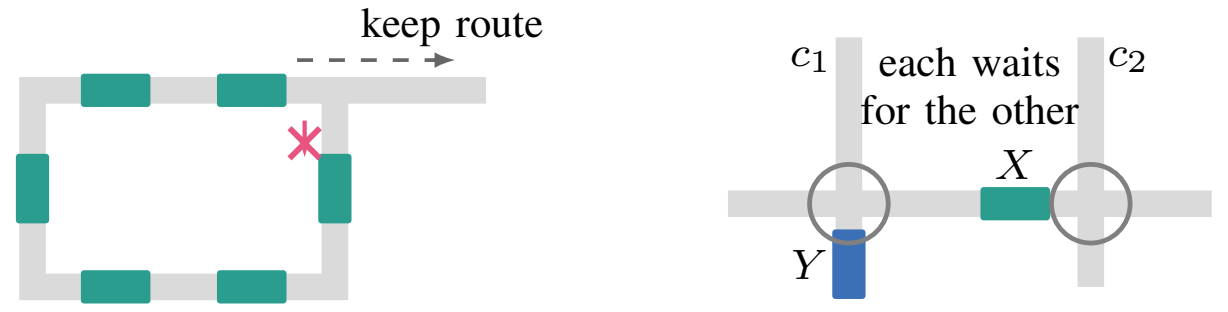

The proof in Appendix~\ref{app:or} goes through the events one by one. A cycle of $\mathcal{W}$ can appear only if a segment
becomes full or the head of a full segment changes its wish, and~\eqref{eq:look} and~\eqref{eq:reroute} exclude exactly these
two cases (Fig.~\ref{fig:route}a). The theorem does not use the planner. A planner that proposes the worst routes changes the throughput, not the invariant. The guard also covers every segment, so a queue it holds back cannot close a ring elsewhere: there is no unguarded ring for the jam to move to.

\subsection{Least intervention}

OpenRing is a safety filter on a discrete state: as a control barrier function filters a nominal input~\cite{ames2017cbf,wabersich2023filter},
the guard filters the proposals of a planner, and, in the spirit of the maximally permissive supervisors of~\cite{ramadge1987supervisory}, among rules that change routes only through proposals every veto of Look and $\Psi$ is one that a rule surviving a planner loss must make. We show that it overrides the planner only
when it must. Call the floor \emph{frozen} if no AGV enters and no route changes, and let $\mathcal{Z}_{\mathrm{clr}}$ be the states from which
some sequence of commitments and exits empties a frozen floor.

\begin{theorem}[Least intervention]\label{thm:least}
Let $B\ge2$. Then
\begin{enumerate}
\item[(i)] $\mathcal{Z}_{\mathrm{clr}}=\mathcal{Z}_{\mathrm{safe}}$;
\item[(ii)] every rule that changes routes only through proposals and satisfies~\eqref{eq:p3} for all $\Pi\in\mathcal{P}$ stays in $\mathcal{Z}_{\mathrm{safe}}$;
\item[(iii)] from $z\in\mathcal{Z}_{\mathrm{safe}}$, a move into a segment with room has $\mathrm{Look}=0$ if and only if it leaves $\mathcal{Z}_{\mathrm{clr}}$, and a route has $\Psi=0$ if and only if it leaves $\mathcal{Z}_{\mathrm{clr}}$.
\end{enumerate}
\end{theorem}

For fixed routes, (i) is the classical maximal permissiveness of the look-ahead. Part (ii) is what the planner adds: since a planner may fall silent at any time, a state outside
$\mathcal{Z}_{\mathrm{safe}}$ cannot be tolerated even if a later route change could repair it. Safety in~\eqref{eq:p1} is not a design choice but a
consequence of clearing, and by (iii) every veto of the
guard is one that any feasible rule changing routes only
through proposals must make.

The remaining conditions of OpenRing are needed as well.

\begin{proposition}[No condition can be dropped]\label{prop:hard}
Each of the following violates~\eqref{eq:problem} for some planner.
\begin{enumerate}
\item[(a)] \emph{No route check.} There are $z\in\mathcal{Z}_{\mathrm{safe}}$ and a route change with $z^{+}\notin\mathcal{Z}_{\mathrm{safe}}$.
\item[(b)] \emph{A binding order, $T_f=\infty$.} There is $z\in\mathcal{Z}_{\mathrm{safe}}$ with $\mathcal{F}(z)=\emptyset$ from which, with entries stopped, no event ever occurs.
\item[(c)] \emph{No limit, $K=\infty$.} An AGV commits forever and never exits.
\item[(d)] \emph{Storage $B=1$.} There is $z\in\mathcal{Z}_{\mathrm{safe}}\setminus\mathcal{Z}_{\mathrm{clr}}$, also on a one-way grid.
\end{enumerate}
\end{proposition}

Fig.~\ref{fig:route}b shows case (b). This is the cycle of a dependency graph known from plan execution~\cite{honig2019persistent,berndt2024receding}. The point here is
that it needs no full segment, so a storage guard alone does not exclude it. Plan execution methods keep the order acyclic
and pay when a vehicle is late. OpenRing lets the order bind for at most $T_f$.

\subsection{Liveness}

\begin{theorem}[Progress]\label{thm:or-live}
Let $B\ge2$, and let $D_0$ bound the time from a commitment until the AGV has cleared its crossing and the next AGV of its segment is a candidate (admitted or not). If AGVs are on the
floor and disturbances have ended, a coordinator that decides at once, or checks with period $\delta\le\min\{T_f,D_0\}$, makes a commitment or an exit within every interval of length
\begin{equation}\label{eq:tbar}
\bar T=T_f+D_0.
\end{equation}
\end{theorem}
Physically, the floor waits at most $T_f$ for a late planned AGV and $D_0$ for the AGVs already committed to finish their crossings; after that, some crossing serves whichever head is ready.

\begin{proposition}[Finite clearing time]\label{prop:d0}
Let the vehicles satisfy (C5). Suppose that on each
internal segment, $B-1$ AGVs queued with the first at
the downstream stop line and successive net gaps
$\gamma_0+\epsilon$ leave a free entry length of at
least $\ell+\gamma_0+\epsilon$, as required by (C1).
Let $\bar d=\max_s d_s$ with $d_s$ from~\eqref{eq:ts},
let $0<\epsilon\le\bar d$, and set
$
T_{\mathrm{seg}}
=\frac{\bar d}{V}+\frac{V}{a_{\max}}
 +t_\gamma\ln\!\frac{\bar d}{\epsilon}.
$
Let $\bar P$ bound the number of AGV bodies on the
entry and internal segments. Then, while no disturbance acts, a committed AGV waits only for AGVs committed before it, so no chain of such waits closes, and $D_0\le(\bar P+1)(\Delta+h_0+B\,T_{\mathrm{seg}})$. The guard alone keeps these waits acyclic in the ledger; (C5) turns this into the time bound. The bound is coarse; measured clearing times stay below $13\,\mathrm{s}$ (Section~\ref{sec:abl}).
\end{proposition}

\begin{corollary}[Clearing]\label{cor:or-clear}
Let $B\ge2$. Define
\begin{equation}\label{eq:phi}
\Phi(z)=\sum_{i\in\mathcal{I}(t)}\bigl(R_i+L(K-\kappa_i)\bigr),
\end{equation}
where $R_i$ is the number of segments ahead of $i$. If no AGV enters after $T_0$ and disturbances have ended, the floor is empty after at most $\Phi(z(T_0))$
commitments and route changes, and $\Phi(z(T_0))\le|\mathcal{I}(T_0)|\,L(K+1)$.
\end{corollary}
Each commitment moves an AGV one segment closer to its exit and each route change spends one of its $K$ credits, so the floor drains like a queue with a finite number of steps left.

Since an order expires, a plan is advice, and the value of advice can be stated from both sides. Call a sequence of commitments a \emph{clean plan} if it follows $\hat\prec_c$
and, along the run, every planned move has $\mathrm{Look}=1$ throughout the time its AGV is a head, every planned route has $\Psi=1$, the commitment before a move at its crossing has been served and the AGV of the move is admitted less than $T_f$ after that commitment, or after $t=0$ for the first commitment at a crossing, with $\tau_c(0)=0$, and each AGV changes route at most $K$ times.

\begin{theorem}[Advice]\label{thm:advice}
Let $B\ge2$ and let disturbances be absent.
\begin{enumerate}
\item[(i)] \emph{A good plan is kept.} Under a clean plan, OpenRing preserves the planned passage order at every crossing: each crossing commits the passages of the plan in the planned order.
\item[(ii)] \emph{A bad plan is bounded.} For every $\Pi\in\mathcal{P}$, if the floor never empties,
\begin{equation}\label{eq:robust}
\Lambda(\mu^\star;\Pi,0)\ \ge\ \frac{1}{L(K+1)\,(T_f+D_0)}.
\end{equation}
\end{enumerate}
\end{theorem}

Thus $T_f$ is the trust placed in the planner. A larger $T_f$ keeps more plans clean and lowers the bound~\eqref{eq:robust}; at $T_f=\infty$ the bound is zero, which is Proposition~\ref{prop:hard}(b). The bound is a worst case, far below the throughput we measure; what it guarantees is that throughput stays positive. In the terms of
algorithms with predictions~\cite{lykouris2021competitive}, with the plan as the prediction, a correct plan keeps its crossing orders and a wrong one cannot stop the floor.

The split of tasks also follows the difficulty of each. Safety costs one walk per move, while the order is what is hard:

\begin{proposition}[Order is the hard part]\label{prop:np}
$\mathrm{Look}$ takes $O(|\mathcal{S}|)$ steps. On general floors, clearing in least time is NP-hard, even with three crossings, unit passages, negligible traversal times, and a guard that never refuses.
\end{proposition}

So a local rule can be exact for safety, and only efficiency needs a planner. Theorem~\ref{thm:or-safe} with Lemma~\ref{lem:commit} gives~\eqref{eq:p1}, Theorem~\ref{thm:or-live} gives~\eqref{eq:p2} with $\bar T$
from~\eqref{eq:tbar}, and Corollary~\ref{cor:or-clear} gives~\eqref{eq:p3}. Hence:

\begin{corollary}[Feasibility]\label{thm:feas}
For $B\ge2$, vehicles that satisfy (C1)--(C5), disturbances that stop only AGVs before their stop line, and segments that meet the length condition of Proposition~\ref{prop:d0}, OpenRing $\mu^\star$ is feasible for~\eqref{eq:problem} for every planner $\Pi\in\mathcal{P}$. If the planner is lost at any time,
$\mu^\star$ continues from that state with $\hat\prec_c$ void and routes frozen, and it remains feasible.
\end{corollary}

\emph{The planner interface.} The guarantees of Theorems~\ref{thm:or-safe}--\ref{thm:or-live} and Corollary~\ref{thm:feas} hold for any planner $\Pi$ that provides the routes and crossing orders specified in~\eqref{eq:planner}, whether it uses a windowed search, a priority rule, or a learned policy. The quality of the planner enters in one place only, the throughput, and Theorem~\ref{thm:advice} bounds how.

\section{The Price of a Guarantee}\label{sec:price}

The objective of~\eqref{eq:problem} has a limit that no rule can pass within a class. A segment that may block at any moment can
accept only what it can hold.

Let $d_s$ be the distance from the stop line in front of $s$ to the stop line at its end, so that
\begin{equation}\label{eq:ts}
T_s=d_s/V
\end{equation}
is the least time between them. Passing a stop line is the point of no return. Let $\mathcal{D}^{\circ}$ be the disturbances that stop, for any time, any AGV
on a segment that has not passed the stop line at its end; an AGV inside a crossing is not stopped by them. A rule is \emph{robust} if, on every
segment and for every $d\in\mathcal{D}^{\circ}$, no AGV ever stops inside a crossing. It may use any information about the past and
the present, including the whole floor. The stops of $\mathcal{D}$ halt uncommitted AGVs, which have not passed their stop line,
so they belong to $\mathcal{D}^{\circ}$. OpenRing is robust: an AGV passes a stop line only with physical room in its target, by (C1)
(Lemma~\ref{lem:commit}), and no disturbance of $\mathcal{D}^{\circ}$ stops it afterwards. The bound below therefore holds for OpenRing. Let $N_s(t)$ count the AGVs past the first stop line of $s$ and not past the second.

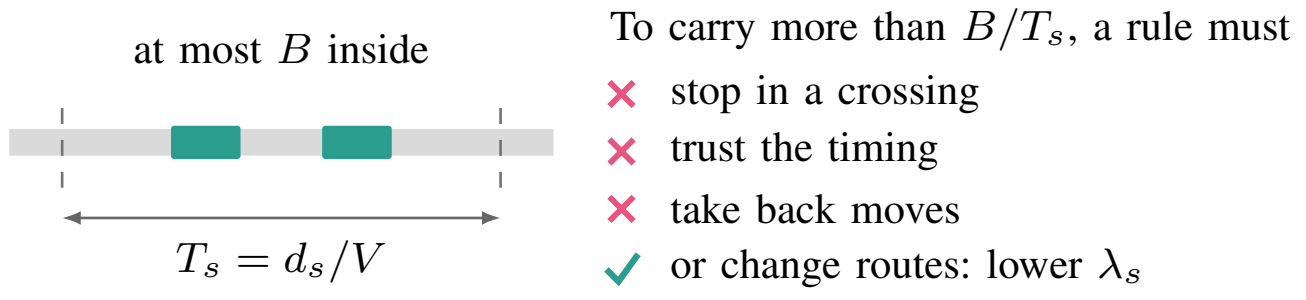
\begin{figure}[t]
\centering
\begin{tikzpicture}[x=1cm,y=1cm,font=\footnotesize]
\begin{scope}[xshift=0cm]
\draw[orroad] (-.15,0) -- (3.45,0);
\draw[black!55,dashed,line width=.5pt] (.2,-.3) -- (.2,.3); \draw[black!55,dashed,line width=.5pt] (3.1,-.3) -- (3.1,.3);
\node[orh] at (1.15,0) {}; \node[orh] at (2.15,0) {};
\node at (1.65,.62) {at most $B$ inside};
\draw[{Latex[length=3.5pt]}-{Latex[length=3.5pt]},black!60,line width=.5pt] (.2,-.5) -- (3.1,-.5);
\node[below] at (1.65,-.52) {$T_s=d_s/V$};
\node[anchor=north,font=\scriptsize] at (1.65,-1.17) {(a) $\lambda_s\le B/T_s$};
\end{scope}
\begin{scope}[xshift=3.7cm]
\node[anchor=west] at (0,.75) {To carry more than $B/T_s$, a rule must};
\foreach \y/\t in {.32/stop in a crossing, -.06/trust the timing, -.44/take back moves} {
  \draw[orRed,line width=1.1pt] (.12,\y-.08) -- (.28,\y+.08) (.12,\y+.08) -- (.28,\y-.08);
  \node[anchor=west] at (.4,\y) {\t};}
\draw[orTeal,line width=1.2pt] (.1,-.84) -- (.18,-.92) -- (.31,-.74);
\node[anchor=west] at (.4,-.83) {or change routes: lower $\lambda_s$};
\node[anchor=north,font=\scriptsize] at (2.0,-1.17) {(b) What a higher flow gives up};
\end{scope}
\end{tikzpicture}%
\vspace{-3mm}\caption{The price of a guarantee. (a) A segment that may block can accept only what it holds. (b) Exceeding the bound gives up a condition; changing routes instead lowers the flow on a segment.}
\label{fig:price}
\end{figure}

\begin{theorem}[Flow bound]\label{thm:flow}
A robust rule satisfies $N_s(t)\le B$ for all $t$, and hence
\begin{equation}\label{eq:flow}
\lambda_s\le\frac{B}{T_s}=\frac{BV}{d_s}.
\end{equation}
\end{theorem}
A segment that may block acts like a lot with $B$ spaces whose cars stay at least $T_s$ each: no rule, however informed, gets more than $B$ cars through it per $T_s$.

The proof, in Appendix~\ref{app:price}, shows $N_s\le B$ for every robust rule and then counts passages as in Little's law; the bound holds for every robust rule, whatever it knows. Here $B$ is the number of stopped AGVs that $s$ can hold
physically; a rule that counts fewer places is bounded by the same number. Fig.~\ref{fig:price} shows both results. Substituting~\eqref{eq:flow} into the load~\eqref{eq:bound}, with $\lambda_s=\eta f_s$ for
a demand $f$ normalized so that $\max_c\eta_c(f)=1$ at load $\eta$, gives an upper bound on the capacity of a floor under a guarantee,
\begin{equation}\label{eq:cap}
\eta\le\eta^{\star}=\min\Bigl\{1,\ \min_{s:\,f_s>0}\frac{B}{T_s f_s}\Bigr\}.
\end{equation}
Bound~\eqref{eq:cap} depends only on the floor, the vehicle and the demand. Appendix~\ref{app:deriv} gives the occupancy integral behind Theorem~\ref{thm:flow} and the circuits of the ring.

\begin{remark}[What a higher flow gives up]\label{cor:giveup}
By Theorem~\ref{thm:flow}, a rule with $\lambda_s>B/T_s$ on some segment is not robust: some stop of $\mathcal{D}^\circ$ leaves an AGV in a crossing with nowhere to go. Such a rule stores AGVs in crossings, relies on AGVs keeping a predicted
timing, or takes back moves past a stop line.
\end{remark}

Bound~\eqref{eq:flow} constrains the flow on a segment, not the demand on the floor: changing routes lowers $f_s$ on loaded segments and
raises $\eta^\star$ without violating anything. The methods of Table~\ref{tab:master} differ in these terms. PIBT in our implementation
stores AGVs in crossings, windowed planning relies on the planned timing, and planners with free routes spread the load. OpenRing
changes routes as well, and its guarantee covers every stop before commitment.

The bound is tight on a lane of the cell model without crossing delay. Take the cell model: a segment is $B$ cells followed by its crossing, an AGV moves one cell per unit step, so $T_s=B+1$, and moves are committed at the stop line. Let $g_j(k)$
be the time at which the $k$-th AGV enters segment $j$ of a saturated lane, and let a free place become known upstream after a delay
$\omega$ of this model, distinct from the late view of Section~\ref{sec:dist}. Then
\begin{subequations}\label{eq:maxplus}
\begin{align}
g_{j+1}(k)&\ge g_j(k)+B+1, \label{eq:mp1}\\
g_j(k)&\ge g_{j+1}(k-B)+\omega, \label{eq:mp2}\\
g_j(k)&\ge g_j(k-1)+1. \label{eq:mp3}
\end{align}
\end{subequations}

\begin{figure*}[t]
\centering
\includegraphics[width=\textwidth]{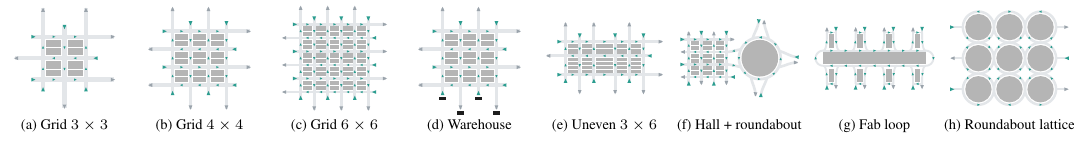}
\caption{The eight floors: one-way aisles with teal entries and grey exits; racks and buildings grey.}
\label{fig:floors}
\end{figure*}

\begin{proposition}[Tightness]\label{prop:tight}
On a saturated lane with at least two internal segments and enough entry capacity, the guard carries the flow
\begin{equation}\label{eq:tight}
\lambda=\frac{B}{B+1+\omega},
\end{equation}
which equals $B/T_s$ in~\eqref{eq:flow} for $\omega=0$.
\end{proposition}

A delay in communication is thus a loss of capacity. A rule that trusts timing is limited on this lane only by~\eqref{eq:mp3}, at one AGV per step, so the guarantee costs exactly the factor $(B+1+\omega)/B$.

A ring shows where the price comes from. Let $Q$ AGVs circulate on a ring of $M\ge2$ segments, with $Q<MB$, since under the guard a ring is never full.

\begin{proposition}[Fundamental diagram of a ring]\label{prop:ring}
With the recursion~\eqref{eq:maxplus} on the ring, the guard carries
\begin{equation}\label{eq:ringfd}
\lambda=\min\Bigl\{\frac{Q}{M(B+1)},\ \frac{MB-Q}{M\,\omega}\Bigr\},
\end{equation}
where the second term is void for $\omega=0$.
\end{proposition}

The two terms are the limits set by the vehicles and by the holes, the structure of the min-plus fundamental diagrams of roads~\cite{farhi2011traffic}, here with storage $B$ and a crossing delay $\omega$. They meet at $Q^\star=MB(B+1)/(B+1+\omega)$, where~\eqref{eq:ringfd} equals the lane flow~\eqref{eq:tight}. Beyond $Q^\star$ the flow falls with the number of holes, to $1/(M\omega)$ at the last one (Fig.~\ref{fig:ring}a). Free places travel against the traffic and
carry the flow, which is why the guard keeps one on every ring.

\section{Experiments}\label{sec:exp}

We test OpenRing at three levels: a closed-loop simulator with full AGV bodies gives the scale, hundreds of runs on eight floors; Isaac Sim checks it with real dynamics and measured poses; and an ideal cell model, last, tests the theory itself.

\subsection{Setup and Metrics}\label{sec:setup}

\begin{table*}[t]
\centering\footnotesize
\begin{threeparttable}
\caption{Methods and what they need, with deadlocks and throughput.}
\label{tab:master}
\setlength{\tabcolsep}{4.6pt}
\begin{tabularx}{\textwidth}{>{\raggedright\arraybackslash}X ccc rr rr r rrr r}
\toprule
\multirow{2}{*}{Method} & \multicolumn{3}{c}{Needs} & \multicolumn{2}{c}{Cell: deadlocked runs} & \multicolumn{2}{c}{Cell: delivered, 0.9} & Delay & \multicolumn{3}{c}{Closed loop} & Isaac\\
\cmidrule(lr){2-4}\cmidrule(lr){5-6}\cmidrule(lr){7-8}\cmidrule(lr){9-9}\cmidrule(lr){10-12}\cmidrule(lr){13-13}
 & route & control & timing & one turn & several & one turn & several & cell & AGVs/min & heavy & deadlocks & deadlocks\\
\midrule
\rowcolor{oursrow} OpenRing & free & central & none & 0/60 & 0/72 & 0.95 & 0.96 & 40 & 49 & 41 & 0/48 & 0/21\\
OpenRing, no planner & fixed & local & none & 0/60 & 0/72 & 0.94 & 0.88 & 201 & 43 & 33 & 0/48 & 0/20\\
First-come & fixed & local & none & 19/60 & 61/72 & 0.50 & 0.11 & -- & 1 & 0 & 36/36 & 13/21\\
PIBT~\cite{okumura2022pibt}, fixed & fixed & central & next step & 38/60 & 31/72 & 0.00 & 0.35 & -- & -- & -- & -- & --\\
PIBT~\cite{okumura2022pibt} & free & central & next step & 0/60 & 0/72 & 0.85 & 0.99 & 5 & 0\rlap{$^\dagger$} & 0\rlap{$^\dagger$} & 12/12\rlap{$^\dagger$} & --\\
LaCAM~\cite{okumura2023lacam} & free & central & window & 0/60 & 0/72 & 0.82 & 0.78 & 812 & 13\rlap{$^\dagger$} & 6\rlap{$^\dagger$} & 10/12\rlap{$^\dagger$} & --\\
RHCR~\cite{li2021lifelong}, fixed & fixed & central & window & 16/60 & 11/72 & 0.41 & 0.60 & 3 & -- & -- & -- & --\\
RHCR~\cite{li2021lifelong} & free & central & window & 4/60 & 1/72 & 0.77 & 0.96 & 3 & 46 & 34 & 6/48 & 0/18\\
SADG~\cite{berndt2024receding} & fixed & central & plan & -- & -- & -- & -- & -- & 26 & 24 & 13/36 & 1/18\\
\bottomrule
\end{tabularx}
\begin{tablenotes}\footnotesize
\item Routes are fixed at entry or may be replanned (free).
Cell-model delivery is the fraction of arrivals delivered at load $0.9$.
Delay is measured in steps at load $0.8$ in deadlock-free runs.
Closed-loop results cover the $3\times3$ and $4\times4$ grids.
Isaac reports deadlocked/total runs for each method, excluding the fab loop.
A dash indicates an untested case or a delay omitted because most runs deadlocked.
$^\dagger$Planner run alone with a binding order at load $0.8$, seeds 0--2.
\end{tablenotes}
\end{threeparttable}
\end{table*}

\begin{figure*}[t]
\centering
\includegraphics[width=\textwidth]{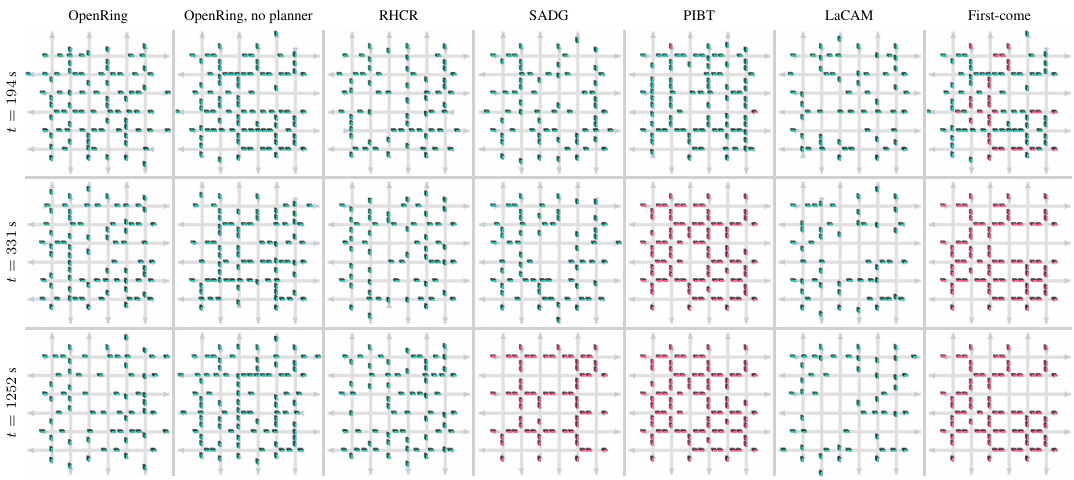}
\caption{Closed loop with full AGV bodies on the $6\times6$ grid at load $0.8$ under heavy disturbance, at three instants. Teal: moving; pink: never moves again.}
\label{fig:snapgrid}
\end{figure*}

\emph{Floors.} Fig.~\ref{fig:floors} shows eight floors: $3\times3$, $4\times4$ and $6\times6$ crossing grids; an uneven $3\times6$ floor with unequal block lengths; and a $4\times4$ warehouse with workstations along one wall, where most routes start or end and $30\%$ of AGVs make one pick stop. Internal segments hold $B=2$ AGVs, and workstations serve as entries and exits. Unless stated otherwise, AGVs arrive as independent streams on multi-turn routes between all entries and exits.
Three floors have curved roads: a $4\times4$ hall opening onto a $10\,\mathrm{m}$-radius roundabout with three two-lane arms; a fab loop with a stadium-shaped main loop and eight bays, as in semiconductor overhead transport; and a $3\times3$ lattice of touching $8\,\mathrm{m}$-radius roundabouts rotating in opposite directions. Lanes meet rings at $55^\circ$; where lanes run close, crossing boxes are lengthened until waiting AGVs clear other lanes, and every internal segment still holds two AGVs. Since the load bound counts these boxes, the hall,
the fab loop and the lattice are tested at $0.8$,
$1.2$ and $1.6$ times the bound, with three seeds
without and under heavy disturbance.

\emph{Vehicles.} The iw.hub body, $1.43\times0.66\,\mathrm{m}$, with bounded acceleration, car following with a time gap and turns in place; collisions are checked on true bodies. In Isaac Sim, PhysX drives the AGVs and the coordinator reads their poses.

\emph{Disturbances.} $20\%$ or $40\%$ of the AGVs stop once in an aisle for $5$--$10\,\mathrm{s}$, and the coordinator may see the floor $0.5\,\mathrm{s}$ late; the heaviest level combines stops of $40\%$ with the delay.

\emph{Methods.} \emph{OpenRing} replans periodically with PBS~\cite{ma2019searching} as in RHCR~\cite{li2021lifelong}, with priority inheritance as in PIBT~\cite{okumura2022pibt}, $T_f=10\,\mathrm{s}$ and $K=10$. Priority inheritance only reorders the candidates $\mathcal{A}_c$, so Theorems~\ref{thm:or-safe}--\ref{thm:or-live} and Corollary~\ref{thm:feas} hold; it reorders $3$--$4\%$ of the commitments, outside Theorem~\ref{thm:advice}(i), and removing it changes the share delivered by under $0.01$ (12 runs each). Planners run on a cell graph, one cell per place and per crossing, with a 40-step window every $2\,\mathrm{s}$ and $1\,\mathrm{s}$ per call; $\hat\prec_c$ ranks AGVs by when their plan first enters $c$, $\hat\rho_i$ follows the planned segments and then a shortest route, and a failed call keeps the last proposal. The vehicles enforce (C1)--(C4) on true positions, so a delayed view affects only the state the coordinator commits on. \emph{OpenRing, no planner} runs the guard alone with fixed routes, the classical look-ahead supervisor~\cite{xing1996deadlock,reveliotis1997polynomial} in FIFO form. \emph{RHCR} is our continuous-time version of~\cite{li2021lifelong}: the same planner, without the guard and the expiry. \emph{ADG}~\cite{honig2019persistent} and \emph{SADG}~\cite{berndt2024receding} execute a plan under a dependency graph, SADG with switching. \emph{First-come} serves arrivals with room, without look-ahead. \emph{PIBT}~\cite{okumura2022pibt} and \emph{LaCAM}~\cite{okumura2023lacam} replace PBS in the same windows, on the directed cell graph without swaps, and run alone with a binding order, as RHCR, or inside OpenRing. All methods meet the same arrivals.

\emph{Runs and metrics.} Loads are fractions of the crossing bound~\eqref{eq:bound}, above the flow bound of Theorem~\ref{thm:flow}: from load $0.6$ the grids saturate, OpenRing delivering about two thirds of the arrivals, so throughput measures attained capacity. Closed-loop runs last at least $1200\,\mathrm{s}$ and physics runs $600\,\mathrm{s}$ unless stated, the first fifth being warm-up. We report throughput, AGVs leaving the floor per minute; deadlocked runs, in which no AGV moves for $60\,\mathrm{s}$; the mean delay $\delta_i=t^{\mathrm{out}}_i-t^{\mathrm{in}}_i-\mathrm{len}(\rho_i)/V$, with arrival and exit times $t^{\mathrm{in}}_i$, $t^{\mathrm{out}}_i$, so that the wait at the entry counts, averaged over the delivered AGVs, and route length $\mathrm{len}(\rho_i)$; body overlaps and the smallest gap between bodies; and plan following, the share of commitments that admit the head PBS ranks first. To detect partial deadlocks, we track every OpenRing AGV: none stands still at the end of its 114 runs. Code will be released upon publication.

\subsection{One Floor, All Methods}\label{sec:main}

Table~\ref{tab:master} and Fig.~\ref{fig:snapgrid} compare the methods on the same arrivals, with stop lines and commitments as in Section~\ref{sec:exec}.

\emph{All methods on one floor.} Fig.~\ref{fig:snapgrid} runs every method on the $6\times6$ grid at load $0.8$ under heavy disturbance. Rings of full segments close around blocks under first-come, under PIBT executed as a binding order and, late in the run, under SADG, until all their AGVs stand still. OpenRing with and without its planner, RHCR and LaCAM keep the floor moving, and OpenRing carries the most, 74 AGVs per minute.

\begin{figure*}[t]
\centering
\includegraphics[width=\textwidth]{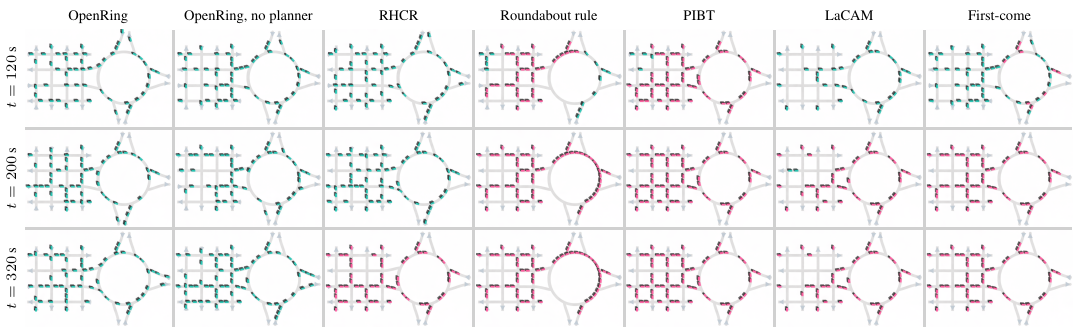}
\caption{Closed loop on the hall with a roundabout at load $0.8$ without disturbance, same arrivals, at three instants, drawn as in Fig.~\ref{fig:snapgrid}. The roundabout rule lets circulating AGVs go first and keeps an AGV whose exit is full waiting on the ring.}
\label{fig:snapcurved}
\end{figure*}

\emph{Deadlock and delay on the $4\times4$ grid.} At load $0.6$ under heavy disturbance, RHCR initially delivers as much as OpenRing. It replans in windows and commits moves without checking whether they close a ring. When stops delay
AGVs, a committed move fills the last free place of a ring, leaving all 36 AGVs stationary from $448\,\mathrm{s}$. These rings of full segments are excluded by Theorem~\ref{thm:least}: with fixed routes, no move order clears them.
SADG maintains its planned order at crossings, even behind an AGV at a pick stop, and delivers $27\%$ fewer AGVs than OpenRing.

\begin{figure}[t]
\centering
\includegraphics[width=\columnwidth]{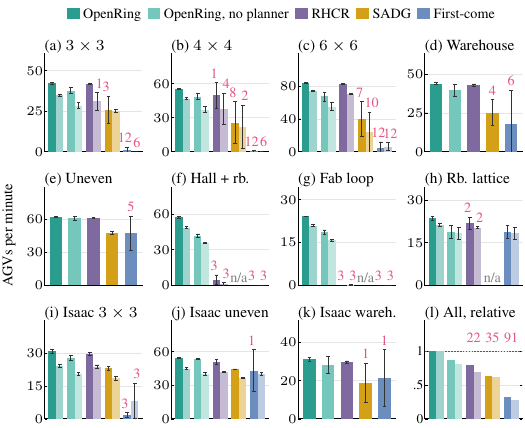}
\caption{All experiments: AGVs per minute, mean $\pm$ s.d. over seeds at the same load (6--12 runs per bar in (a)--(c), 9 in (d), (e), 3 in (f)--(k)); dark: no disturbance, light: heavy; pink: deadlocked runs. (l) Throughput relative to OpenRing, averaged over (a)--(k), and all deadlocked runs.}
\label{fig:summary}
\end{figure}

\emph{A binding order stops the floor with room left.} In a run on the $6\times6$ grid at load $0.6$ under heavy disturbance, SADG fails in its second way. When the AGV a crossing has planned next is late or held by an aisle stop, the crossing keeps its order: the next AGV waits at its stop line with its target free, and the waits pass upstream through the aisles. All 72 AGVs stand still by $248\,\mathrm{s}$. This is the
binding order of Proposition~\ref{prop:hard}(b): the floor stops while the planned AGV is late, with room left on it. No segment need be full, so counting free places cannot detect this deadlock; hence OpenRing's orders expire.

\emph{OpenRing keeps the floor moving.} With the same planner as RHCR, in the run of Fig.~\ref{fig:snapgrid} at most three of its AGVs stand for more than $20\,\mathrm{s}$ at any time. Over the run the planner changes thousands of routes, each one checked by $\Psi$, and the order lapses hundreds of times, each time passing an AGV that a binding order would have held behind a late one.

\subsection{Curved Floors}\label{sec:curved}

\emph{Roundabouts and loops need no new rule.} The guard reads chains of FIFO segments and nothing of the grid (Proposition~\ref{thm:look}), so OpenRing runs unchanged on the three curved floors (Fig.~\ref{fig:snapcurved}): in its 18 runs there it has no deadlock or overlap and carries the most on each floor (Fig.~\ref{fig:summary}). On the hall and on the fab loop every run of RHCR and first-come deadlocks, and so does every run of the usual roundabout rule: by letting circulating AGVs go first, it keeps an AGV whose exit is full waiting on the ring, until a ring of full segments closes through the roundabout and the hall. A closed route can carry traffic as long as the AGVs on it
have room to advance. Look intervenes when a proposed move
would fill the last place and leave each head in the cycle
waiting for the next full segment (Lemma~\ref{lem:inv}). On the roundabout lattice the guard refuses no move at these loads, so OpenRing without a planner runs as first-come and neither deadlocks, while RHCR deadlocks in four of six runs.

\begin{figure}[t]
\centering
\includegraphics[width=\columnwidth]{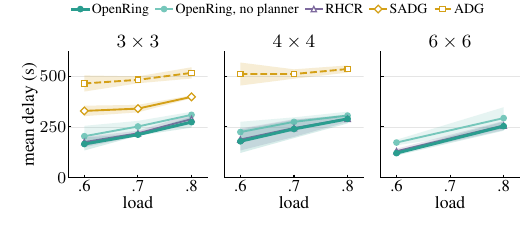}
\caption{Mean delay $\delta_i$ (Section~\ref{sec:setup}) against the load on three grids (loads $0.6$ and $0.8$ on the $6\times6$ grid), closed loop: mean over runs without deadlock, shade $\pm$1 s.d.; first-come deadlocks.}
\label{fig:cl}
\end{figure}

\begin{figure*}[t]
\centering
\includegraphics[width=\textwidth]{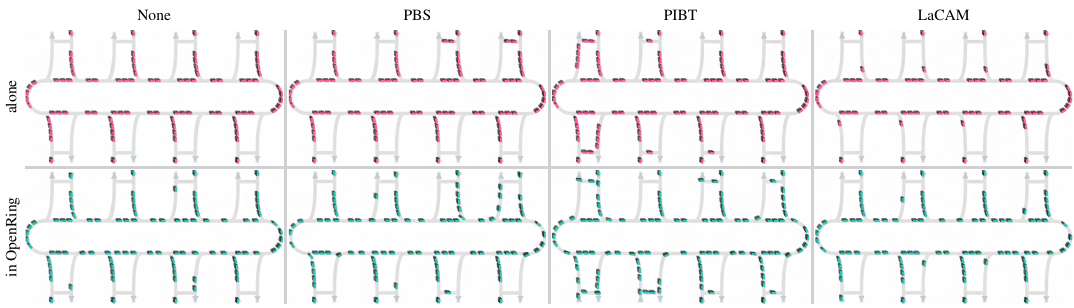}
\caption{The fab loop in Isaac Sim at load $1.2$ without disturbance, same arrivals, one run each, at $300\,\mathrm{s}$ from the logged poses: no planner, PBS, PIBT and LaCAM alone with a binding order (top) and inside OpenRing (bottom); a deadlocked run is shown $60\,\mathrm{s}$ after the deadlock. Colours as in Fig.~\ref{fig:snapgrid}.}
\label{fig:plug}
\end{figure*}

\subsection{All Floors}\label{sec:allfloors}

\emph{OpenRing never deadlocks and carries at least as much.} Over 114 runs on the $3\times3$, $4\times4$ and $6\times6$ grids, at loads $0.6$ to $0.8$ without and under the disturbances of Section~\ref{sec:dist}, OpenRing has no deadlock and no body overlap. Without disturbances it carries 49 AGVs per minute on the two smaller grids against 46 for RHCR, and more in 20 of 24 paired runs (Wilcoxon $p=1.4\times10^{-4}$), while RHCR and SADG deadlock in 1 and 11 of 24 runs. Removing the guard alone changes no run here (Section~\ref{sec:abl}): the margin comes from executing the order, since the expiry $T_f$ and the guard each release AGVs that an inconsistent order holds, and either one alone keeps the margin. SADG deadlocks even without disturbances. Its switch keeps each pair consistent, as in~\cite{berndt2024receding}, whose cells hold one AGV, but on segments of $B$ places it can close a chain of waits through third AGVs and full segments, which only a check on the counts, the check of Look, excludes; planned with up to twice the room, it still deadlocks. ADG never switches and never deadlocks here, but it waits for every late AGV and carries the least of the planned methods, $16$ to $18$ AGVs per minute on the two smaller grids.

\emph{Throughput and deadlocks across layouts.} Across the layouts in Fig.~\ref{fig:summary}, OpenRing achieved throughput comparable to or higher than the best baseline for each layout. We observed no deadlocks in any OpenRing run, with or without the planner. From no disturbance to the heaviest disturbance, throughput on the grids fell by up to $17\%$ for OpenRing and $26\%$ for RHCR. Under heavy disturbance on the $6\times6$ grid, SADG deadlocked in 10 of 12 runs. Without its planner, OpenRing delivered $10$--$26\%$ fewer AGVs on the grids. On the uneven and warehouse floors, RHCR stayed close to OpenRing in throughput, while SADG and first-come deadlocked in some runs. The runs with continuing arrivals show how much traffic the
floor carries through a disturbance. Once disturbances end,
the floor is guaranteed to make progress, and if new entries
also stop, every remaining AGV eventually exits
(Theorem~\ref{thm:or-live} and Corollary~\ref{cor:or-clear}).

\emph{Load and delay.} From load 0.6 to 0.8 the throughput of OpenRing stays flat on each grid, while RHCR follows it until its runs deadlock and then falls behind. Over the runs without deadlock, OpenRing has the lowest mean delay at every point (Fig.~\ref{fig:cl}), tied with RHCR on the $4\times4$ grid at load 0.8. RHCR adds at most $15\,\mathrm{s}$ and OpenRing without a planner up to a minute, while plan execution waits behind every late AGV and adds minutes, ADG up to $331\,\mathrm{s}$. SADG is not shown on the larger grids, where most of its runs deadlock.

\begin{figure}[t]
\centering
\includegraphics[width=\columnwidth]{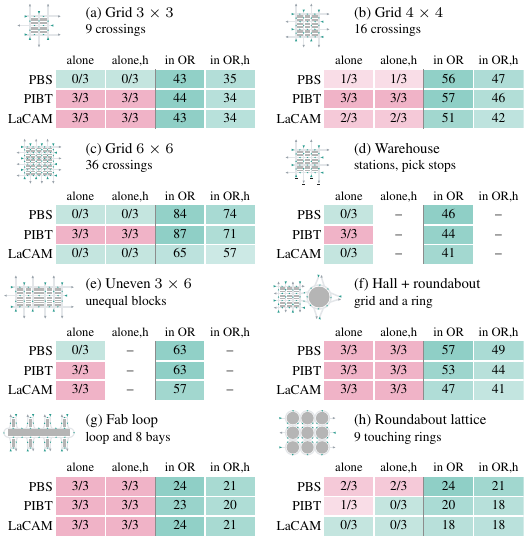}
\caption{Planners alone and inside OpenRing (OR) on the eight floors: deadlocked runs of three executed alone with a binding order (teal: none, pink: some) and AGVs per minute inside OpenRing (teal); h: heavy disturbance.}
\label{fig:plugcards}
\end{figure}

\begin{figure*}[t]
\centering
\includegraphics[width=\textwidth]{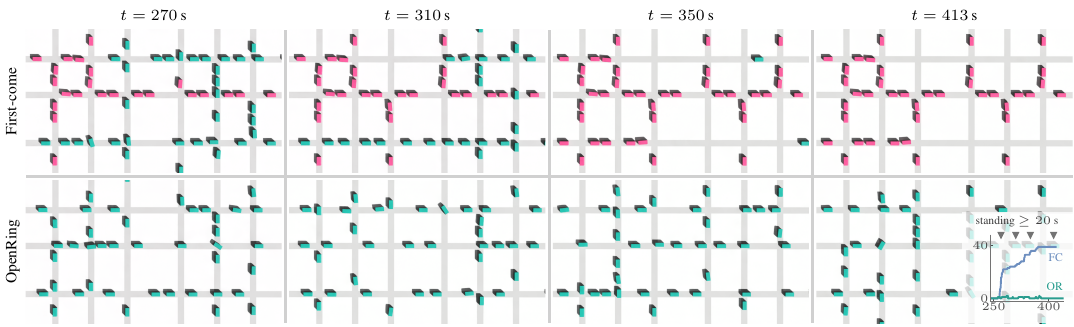}
\caption{Isaac Sim on the uneven $3\times6$ floor, same arrivals: first-come (top) and OpenRing (bottom) at four instants, from the logged poses; inset: AGVs standing for $20\,\mathrm{s}$ or more under first-come (FC) and OpenRing (OR), triangles at the four instants. Colours as in Fig.~\ref{fig:snapgrid}.}
\label{fig:cosim}
\end{figure*}

\begin{figure}[t]
\centering
\includegraphics[width=\columnwidth]{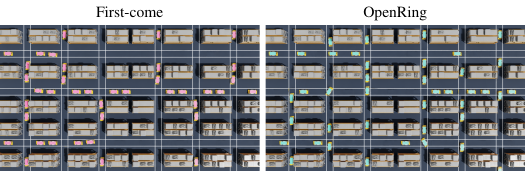}
\caption{The run of Fig.~\ref{fig:cosim} in Isaac Sim at $t=413\,\mathrm{s}$: first-come (left) and OpenRing (right). Pink: AGVs that never move again.}
\label{fig:cosimisaac}
\end{figure}

\subsection{Any Planner Plugs In}\label{sec:plug}

\emph{Any planner plugs in.} With the simplest planner we know, a rank by entry time at every crossing on fixed routes, the plan executed as a binding order, without the guard and the expiry, deadlocks in 23 of 24 runs on the $3\times3$ and $4\times4$ grids. Inside OpenRing it deadlocks in none and delivers $48\%$ and $43\%$ of the arrivals without and under heavy disturbance, against $67\%$ and $56\%$ with PBS.

\emph{Planners from the MAPF literature plug in too.} We replace PBS by PIBT~\cite{okumura2022pibt} and LaCAM~\cite{okumura2023lacam} on every floor of Fig.~\ref{fig:summary} (Figs.~\ref{fig:plug} and~\ref{fig:plugcards}), seeds 0--2, at load 0.8 on the grid and aisle floors and at one load on each curved floor ($0.8$, $1.2$ and $1.6$ on the hall, fab loop and lattice), without and, except on the warehouse and uneven floors, under heavy disturbance. Executed alone, a planner's order binds as in RHCR, so PBS alone is RHCR. Alone, PBS deadlocks in 18 of 42 runs, PIBT in 37 and LaCAM in 25; inside OpenRing none of the 126 runs deadlocks and no bodies overlap. The guard leaves the throughput to the planner: on the $6\times6$ grid PBS and PIBT add about 30\% to OpenRing without a planner, while LaCAM, which spends most of its time limit on each call, adds nothing.

\emph{Over a thousand AGVs.} In one run each, under different conditions, PIBT executed alone deadlocks at $646\,\mathrm{s}$ with 1009 AGVs on a $26\times26$ grid at load $0.6$. Inside OpenRing, on a $31\times31$ grid at load $0.9$, 3100 AGVs enter and up to 1108 are on the floor at once, with no deadlock or overlap; steps average $8.1\,\mathrm{ms}$ and planner calls $30\,\mathrm{ms}$. The supplementary video shows both runs.

\emph{Computation.} On one AMD Ryzen~9 9950X3D core, coordinator steps take $0.8\,\mathrm{ms}$ median and PBS calls at most $5\,\mathrm{ms}$. On grids up to $12\times12$ with 216 AGVs, OpenRing has no deadlock or overlap, with steps of $7.4\,\mathrm{ms}$ on average.

\subsection{Physics Simulation}\label{sec:isaac}

\emph{Rings close in physics as in the model.} On the uneven floor, under first-come the first AGV stops for good at $243\,\mathrm{s}$ and others follow within seconds, as rings of full segments close around the blocks, until all 39 AGVs stand still and deliveries end. OpenRing, on the same arrivals and the same physics, never has more than two AGVs standing for $20\,\mathrm{s}$ and keeps delivering. On the $6\times6$ grid it delivers 67 AGVs per minute, while under first-come a ring of four full segments around one block closes in the logged poses and soon every AGV on the floor stands still. The supplementary video shows both floors.

\emph{Body clearance in Isaac Sim.}
Across the 98 runs of the five methods in Table~\ref{tab:master},
AGV bodies remained at least $16\,\mathrm{cm}$ apart, including
during braking and in-place turns. OpenRing has no deadlock in
any of its 21 runs and carries the most on every physics floor
of Fig.~\ref{fig:summary}, tied with RHCR on the $3\times3$ grid
under heavy disturbance, while first-come deadlocks in 13 of
its 21 runs. Under aisle stops and a delayed view of the floor, OpenRing delivers more than RHCR in all six paired runs, by up to $8\%$ on the uneven floor. On the warehouse floor, where pick stops hold AGVs in the aisles, the methods keep the order of the closed loop. On the hall the usual roundabout rule deadlocks and OpenRing does not; on the fab loop (Fig.~\ref{fig:plug}) every planner executed alone deadlocks within $250\,\mathrm{s}$, and inside OpenRing none does (Figs.~\ref{fig:cosim} and~\ref{fig:cosimisaac}).

\emph{The floor survives the loss of the planner.} In a physics run on the $3\times3$ grid the planner is switched off at $200\,\mathrm{s}$ of $600\,\mathrm{s}$; the floor keeps moving without deadlock or overlap (Corollary~\ref{thm:feas}) and delivers 29 AGVs per minute against 31 with the planner, near the 28 of OpenRing without a planner.

\begin{figure}[t]
\centering
\includegraphics[width=\columnwidth]{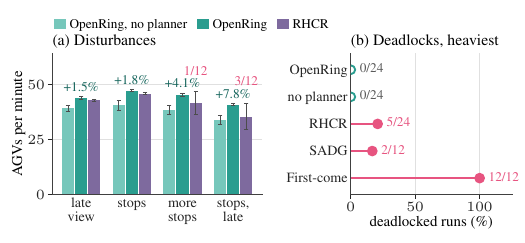}
\caption{(a) Disturbances, mean $\pm$ s.d. over seeds; above the OpenRing bar its gain over RHCR in paired runs, in \% of the arrivals; above the RHCR bar its deadlocked runs (pink); (b) deadlocked runs under the heaviest disturbance.}
\label{fig:global}
\end{figure}

\subsection{Disturbances}\label{sec:dist}

\emph{OpenRing gains most where plans fail.} Over 48 paired runs on the $3\times3$ and $4\times4$ grids at the four disturbance levels (Fig.~\ref{fig:global}a; two loads, seeds 0--2), OpenRing delivers more than RHCR in 44 (sign test $p<10^{-8}$), with no deadlock against 4 for RHCR. The gain grows with the disturbance, from $1.5\%$ of the arrivals under the late view alone to $7.8\%$ under stops with the late view, where OpenRing delivers more in all 12 pairs: a late AGV breaks the timing a plan assumes, and RHCR keeps executing the plan while OpenRing lets the order lapse after $T_f$. Under the heaviest disturbance only OpenRing, with or without its planner, stays free of deadlock over six seeds (three for SADG and first-come), while first-come deadlocks in every run (Fig.~\ref{fig:global}b).

\subsection{Ablation}\label{sec:abl}

\begin{figure}[t]
\centering
\includegraphics[width=\columnwidth]{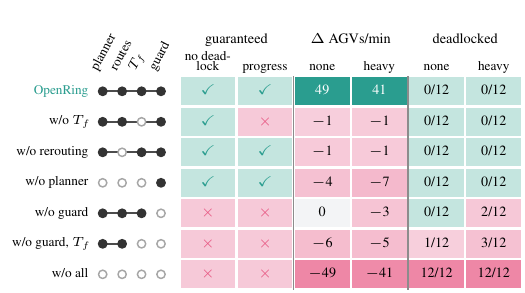}
\caption{Ablation: parts on (filled) or off (open), the guarantees that remain for every planner (Theorems~\ref{thm:or-safe} and~\ref{thm:or-live}), AGVs per minute against OpenRing ($\Delta$; OpenRing row absolute) and deadlocked runs, without and under heavy disturbance. The last two rows execute as RHCR and first-come; the runs are a subset of those of Table~\ref{tab:master}.}
\label{fig:abl}
\end{figure}

Fig.~\ref{fig:abl} removes one part at a time on the $3\times3$ and $4\times4$ grids, 12 runs per condition. \emph{The parts protect each other.} Without a planner the guard is the only protection: with it no run deadlocks, without it every run does. With a planner, removing the guard alone changes nothing without disturbances but lets heavy disturbance close rings, and removing it together with the expiry gives RHCR, which carries less and deadlocks. The planner supplies most of the throughput.

\emph{The plan stays in charge.} $94\%$ of the commitments follow the plan without and $89\%$ under heavy disturbance (Theorem~\ref{thm:advice}); priority inheritance gives $3$--$4\%$, expired orders $2$--$5\%$ and guard refusals under $2\%$.

\emph{How long to trust the plan.} In a sweep of $T_f$ over $84$ runs, throughput peaks when the order lapses after $5\,\mathrm{s}$ without and $10\,\mathrm{s}$ under heavy disturbance, $6\%$ and $12\%$ above $T_f=0$: a shorter expiry discards orders that are still right, a longer one keeps AGVs waiting behind a late one. With $T_f=\infty$ the floor still moves here because PBS replans every period; the expiry guarantees progress for every planner.

\emph{The safeguards rarely act.} At most one committed AGV waits before each stop line, every commitment clears its crossing within $13\,\mathrm{s}$ without disturbance and $25.3\,\mathrm{s}$ under heavy disturbance, the longest observed counterparts of $D_0$ in Theorem~\ref{thm:or-live}, and $\Psi$ refuses one of about 76{,}000 proposed route changes.

\emph{More places per segment.} With $B=3$ and $8\,\mathrm{m}$ segments, none of $72$ runs deadlocks on these grids, and OpenRing still carries more than RHCR in $22$ of $24$ paired runs.

\subsection{The Ideal Cell Model}\label{sec:ideal}

\emph{The ideal model is the home ground of planners.} Execution is exact, vehicles have no bodies, and an AGV may wait inside a crossing. PIBT exploits this by storing AGVs in crossings and RHCR by trusting exact timing; with routes of several turns both deliver slightly more than OpenRing, which uses neither and never deadlocks while RHCR does ($0.998$, $0.991$ and $0.989$ of the arrivals over 72 runs at loads $0.6$ to $0.9$); with one turn per route OpenRing delivers the most (Table~\ref{tab:master}). LaCAM never deadlocks in 132 runs, but its delay reaches 812 steps on multi-turn routes at load $0.8$.
It replans every step and finds a solution within $1\,\mathrm{s}$ on 46\% of calls for one-turn routes and 36\% for multi-turn routes, keeping the previous plan otherwise. The guarantee costs under one percent of the deliveries where it is least needed, and it shows as waiting at stop lines, which exact timing makes free for the others; once bodies and stops make timing slip, OpenRing has the lowest delay over the runs without deadlock (Fig.~\ref{fig:cl}).

\begin{figure}[t]
\centering
\includegraphics[width=\columnwidth]{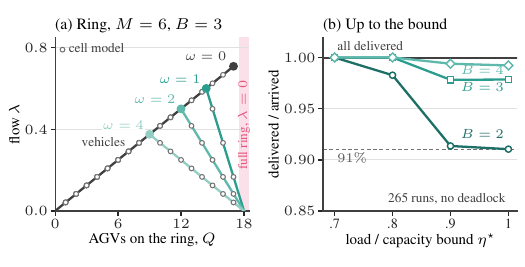}
\caption{Theory against the cell model. (a) Fundamental diagram of a ring: Proposition~\ref{prop:ring} (lines: the vehicle limit, shared by every delay $\omega$, and the hole limit for each $\omega$; capacity filled) and the cell model (circles); a full ring has zero flow, a state the guard never allows. (b) OpenRing without a planner on grids up to the capacity bound $\eta^\star$.}
\label{fig:ring}
\end{figure}

\emph{The model confirms the theory.} On a ring the simulated flow matches Proposition~\ref{prop:ring} at every point, including the delay $\omega$ (Fig.~\ref{fig:ring}a), and on grids OpenRing without a planner delivers at least nine in ten arrivals up to the capacity bound $\eta^\star$, without a deadlock (Fig.~\ref{fig:ring}b).

\section{Discussion}\label{sec:disc}

\emph{Faults before scale.} We studied late and stopped AGVs,
delayed observations and planner loss. Each decision of the guard reads one wait chain, so its cost grows with the full segments on that chain, not with the floor; floors of ten thousand AGVs, where search-based planners are evaluated, are left to future work, as is an asynchronous guard that removes the single coordinator.

\emph{Where the guarantees stop.} They need every internal segment to hold two AGVs. With one place, an AGV that enters fills its segment, so the hole the guard keeps must be a whole empty segment, and a safe state need not be clearable (Proposition~\ref{prop:hard}(d)); single-cell floors need a coarser guard. Two-way aisles with a lane per direction are pairs of one-way segments, and lane changes at fixed points are crossings; a single lane shared by both directions is outside the model. The guarantees hold for the floor, not for each AGV: a bound on every wait needs a fair rule after the order expires.

\section{Conclusion}\label{sec:concl}

Planning and safety need not be one problem: the planner decides how fast the floor runs, and the guard decides that it keeps running. OpenRing lets a planner propose freely and keeps one invariant locally: no ring of full segments. The invariant holds under any proposal and after the loss of the planner; the order of a plan stays advice, because a binding order can stop the floor with room left; and among rules that change routes only through proposals, the guard intervenes only where every feasible rule must. The guarantee has a stated price, at most $B/T_s$ per segment under arbitrary stops. In closed loop and in physics simulation the floor had no deadlock, and in closed loop it carried about as much as the best baseline or more. Open directions are the exact capacity of a network under a guarantee, an asynchronous OpenRing, a bound on the wait of every AGV, and planners aware of holes.

\appendices
\section{Proofs}\label{app:proofs}


Proofs refer to the counted state~\eqref{eq:zstate}. A move $s\to s'$ changes only $n_s$ (minus one), $n_{s'}$ (plus one), the head
of $s$, and the head of $s'$ if $s'$ was empty. Entry segments are never targets of moves and exit segments are never full.

\subsection{Local guard}\label{app:guard}

\begin{lemma}[Per-crossing bound]\label{lem:bound}
If the queues in front of a crossing $c$ stay bounded, then $\eta_c\le1$ in~\eqref{eq:bound}.
\end{lemma}
\begin{proof}
Let $\lambda_a\ge\lambda_b$. On an interval of length $\tau$ the crossing serves $\lambda_a\tau$ and $\lambda_b\tau$ AGVs up to $o(\tau)$, of which at least
$(\lambda_a-\lambda_b)\tau$ consecutive pairs share a segment and need $h_0$. The busy time obeys
\begin{equation}
(\lambda_a+\lambda_b)\tau\Delta+(h_0-\Delta)(\lambda_a-\lambda_b)\tau+o(\tau)\le\tau,
\end{equation}
which gives $\eta_c\le1$ as $\tau\to\infty$.
\end{proof}

\begin{lemma}[A head can move]\label{lem:stuck}
If $\mathcal{I}(t)\ne\emptyset$ and $z\in\mathcal{Z}_{\mathrm{safe}}$, then there is a head with $\nx(s)\notin\mathcal{F}(z)$.
\end{lemma}
\begin{proof}
Suppose $\nx(s)\in\mathcal{F}(z)$ for every non-empty $s$. From a non-empty $y_0$ define
\begin{equation}
y_{k+1}=\nx(y_k),\qquad k\ge0.
\end{equation}
Then $y_k\in\mathcal{F}(z)$ for $k\ge1$, so the sequence is defined for all $k$. Since $|\mathcal{S}|<\infty$, there are $k_1<k_2$ with
$y_{k_1}=y_{k_2}$, $k_1\ge1$, and $(y_k,y_{k+1})\in\mathcal{W}(z)$ by~\eqref{eq:wait}. This is a cycle, against~\eqref{eq:safe}.
\end{proof}

Hence every deadlock contains a full wait cycle.

\begin{lemma}[Commitment keeps safety]\label{lem:inv}
If $z\in\mathcal{Z}_{\mathrm{safe}}$ and $\mathrm{Look}(z,s\to s')=1$, then $z'\in\mathcal{Z}_{\mathrm{safe}}$.
\end{lemma}
\begin{proof}
A move gives $n'_s=n_s-1<B$ and $n'_{s'}=n_{s'}+1$, and leaves every other segment unchanged, so
\begin{equation}\label{eq:dW}
\mathcal{F}(z')\subseteq\mathcal{F}(z)\cup\{s'\},\qquad s\notin\mathcal{F}(z').
\end{equation}
By~\eqref{eq:wait} and~\eqref{eq:dW}, every edge of $\mathcal{W}(z')$ not in $\mathcal{W}(z)$ has $s'$ as an endpoint. Since
$\mathcal{W}(z)$ is acyclic, a cycle of $\mathcal{W}(z')$ passes through $s'$. As $\nx'$ is a function, the cycle is the chain~\eqref{eq:walk}, so
$\chi(z')=1$ and $\mathrm{Look}=0$ by~\eqref{eq:look}, a contradiction.
\end{proof}

\begin{lemma}[An alternative move]\label{lem:alt}
Let $B\ge2$. If $\mathrm{Look}(z,s\to s')=0$ and $n_{s'}<B$, there is $u\ne s$ with $\mathrm{dn}(u)=\mathrm{up}(s')$, $\nx(u)=s'$ and
$\mathrm{Look}(z,u\to s')=1$.
\end{lemma}
\begin{proof}
By~\eqref{eq:look}, $\chi(z')=1$, so in $z'$
\begin{equation}\label{eq:chain}
s'=x_0\to x_1\to\!\dots\!\to x_k=s', x_0,\dots,x_{k-1}\in\mathcal{F}(z'),
\end{equation}
with $k$ the first return to $s'$ ($k\ge2$, since no segment follows itself on a route). Since $s'\in\mathcal{F}(z')$ and $n_{s'}<B$ in $z$, $n_{s'}=B-1\ge1$, the head of $s'$ is the same in $z$ and $z'$, so $x_1$ does not depend on the moving AGV. Let $u=x_{k-1}$. Then $\nx(u)=s'$ and
$\mathrm{dn}(u)=\mathrm{up}(s')$ since $s'$ has one feeding crossing, and $u\ne s$ by~\eqref{eq:dW}. Apply $u\to s'$ to $z$. Now $n_u<B$, so the chain~\eqref{eq:chain}
stops at $u$ and $\chi=0$, which gives $\mathrm{Look}(z,u\to s')=1$.
\end{proof}

\begin{proof}[Proof of Proposition~\ref{thm:look}]
(i) is Lemma~\ref{lem:inv}. (ii) By (i), Lemma~\ref{lem:stuck} gives a head with room ahead or an exit; Lemma~\ref{lem:alt} then gives
\begin{equation}\label{eq:nonempty}
\exists\,c\in\mathcal{C}:\quad \mathcal{A}_c(z)\ne\emptyset,
\end{equation}
and a work-conserving crossing serves an element of $\mathcal{A}_c$ within one separation $h_0$.
\end{proof}

\begin{proof}[Proof of Lemma~\ref{lem:commit}]
Events are applied sequentially, and each commitment passes
$\mathrm{Look}$ on the ledger itself, so Lemma~\ref{lem:inv}
preserves ledger safety regardless of observation delays. For the bodies, let $P_s(t)$ be the AGVs physically on $s$. An AGV in $P_s$ is counted in
$s$, or it is committed out of $s$ and has not yet passed its stop line. A body enters $s$ only through a passage, which by (C1)
starts when the entry of $s$ has room; inside $s$, (C2) keeps the gap to the AGV ahead; through a crossing, (C3) separates the
passages that share a segment and (C4) those of different movements, including a rotation in place and the AGVs waiting at the other stop lines. Hence~\eqref{eq:body} holds. A disturbance may stop an AGV only before its stop line, so a stopped AGV stays outside the crossing and the ledger and the argument are unchanged. Once an AGV passes its stop line, (C1) gives its
target room for it, so it leaves the crossing without being held.
\end{proof}

\subsection{OpenRing}\label{app:or}

\begin{proof}[Proof of Theorem~\ref{thm:or-safe}]
By~\eqref{eq:wait}, $\mathcal{W}$ depends on $z$ only through $\mathcal{F}$ and through $\nx$ on $\mathcal{F}$. So the proof tracks these two objects. Two facts hold for every event.

\emph{Heads of full segments.} The head of $x$ changes only if the head leaves $x$ by a commitment or an exit, or if $x$ was empty. In both cases $x\notin\mathcal{F}$ right after the event, by~\eqref{eq:dW}. So the head of a full segment never changes while it stays full. A segment that becomes full again does so through a commitment into it, which is tested on the wish of its head at that moment.

\emph{Counted AGVs.} A committed AGV is counted in its target by~\eqref{eq:occ}. Its uncommitted route starts there, so a route change acts on the head of the segment that counts it, and an AGV between two segments needs no separate case.

Let $z\in\mathcal{Z}_{\mathrm{safe}}$ and $z^{+}=\Gamma(z,e)$.
\begin{enumerate}
\item $e=\mathsf{commit}$: $\mathrm{Look}=1$ by~\eqref{eq:admit}, so $z^{+}\in\mathcal{Z}_{\mathrm{safe}}$ by Lemma~\ref{lem:inv}.
\item $e=\mathsf{enter}$: only an entry segment changes, and $\mathcal{F}\subseteq\mathcal{S}_{\mathrm{int}}$, so $\mathcal{W}(z^{+})=\mathcal{W}(z)$.
\item $e=\mathsf{exit}$: $\mathcal{F}(z^{+})\subseteq\mathcal{F}(z)$, and the segment left is not full, so $\mathcal{W}(z^{+})\subseteq\mathcal{W}(z)$.
\item $e=\mathsf{reroute}$ of $i$, with $i$ counted in $x$:
\begin{enumerate}
\item $i\ne h(x)$. No wish changes, so $\mathcal{W}(z^{+})=\mathcal{W}(z)$. If $i$ later becomes the head, then $x$ has lost an AGV, and the first fact applies.
\item $i=h(x)$ and $x\notin\mathcal{F}$. Then $x$ has no outgoing edge in~\eqref{eq:wait}, so $\mathcal{W}(z^{+})=\mathcal{W}(z)$.
\item $i=h(x)$ and $x\in\mathcal{F}$. The change is accepted only with $\Psi=1$, so $z^{+}=z[\hat\rho_i]\in\mathcal{Z}_{\mathrm{safe}}$ by~\eqref{eq:reroute}.
\end{enumerate}
\end{enumerate}
Selection~\eqref{eq:select}, $\tau_c$ and $\kappa_i$ do not enter~\eqref{eq:wait}. This proves~\eqref{eq:inv}.
\end{proof}

\begin{proof}[Proof of Theorem~\ref{thm:least}]
(i) Let $z\notin\mathcal{Z}_{\mathrm{safe}}$, with a cycle $u_0\to\dots\to u_k=u_0$ of $\mathcal{W}(z)$. On a frozen floor, suppose some AGV leaves a segment of the cycle, and take the first such event, from $u_l$. Its head wishes $u_{l+1}$, which has lost no AGV and is still full, so no move is possible. Thus $z\notin\mathcal{Z}_{\mathrm{clr}}$, giving
$\mathcal{Z}_{\mathrm{clr}}\subseteq\mathcal{Z}_{\mathrm{safe}}$.

Let $z\in\mathcal{Z}_{\mathrm{safe}}$. By Lemmas~\ref{lem:stuck} and~\ref{lem:alt}, a non-empty floor has an exit or a move with $\mathrm{Look}=1$. By Lemma~\ref{lem:inv} the state stays in $\mathcal{Z}_{\mathrm{safe}}$, and each move lowers $\sum_iR_i$ by one. So the floor empties and $z\in\mathcal{Z}_{\mathrm{clr}}$.

(ii) Let $z$ be reachable under $\mu$ with some $\Pi$. Let $\Pi'$ agree with $\Pi$ until then and stay silent afterwards, and let entries stop there. Routes change only through proposals, so the floor is frozen. Constraint~\eqref{eq:p3} gives $z\in\mathcal{Z}_{\mathrm{clr}}=\mathcal{Z}_{\mathrm{safe}}$ by (i).

(iii) Let $n_{s'}<B$. By~\eqref{eq:look}, $\mathrm{Look}=0$ means $\chi(z')=1$, which is $z'\notin\mathcal{Z}_{\mathrm{safe}}$ by~\eqref{eq:chain}. Conversely, $\mathrm{Look}=1$ gives $z'\in\mathcal{Z}_{\mathrm{safe}}$ by Lemma~\ref{lem:inv}. For routes, the cases of~\eqref{eq:reroute} with $\Psi=1$ have $z[\hat\rho_i]\in\mathcal{Z}_{\mathrm{safe}}$ by the proof of Theorem~\ref{thm:or-safe}, and $\Psi=0$ means $z[\hat\rho_i]\notin\mathcal{Z}_{\mathrm{safe}}$. Part (i) translates both.
\end{proof}

\begin{proof}[Proof of Proposition~\ref{prop:hard}]
(a) Let $x_0\to x_1\to\dots\to x_{m-1}\to x_0$ be a directed cycle of full segments, where the head of $x_l$ wishes $x_{l+1}$ for $l=1,\dots,m-1$ and the head of $x_0$ wishes a segment with room. Then $\mathcal{W}(z)$ is a path, and $z\in\mathcal{Z}_{\mathrm{safe}}$. A planner that sends $h(x_0)$ to $x_1$ closes the cycle.

(b) Let $X$ be the head of a segment into $c_2$ and $Y$ of a segment into $c_1$, with $c_2$ before $c_1$ on $\rho_X$, $c_1$ before $c_2$ on $\rho_Y$, $\mathcal{F}(z)=\emptyset$, and let $X$ and $Y$ be the only AGVs on the floor, with entries stopped. Let $\Pi$ propose
\begin{equation}
X\,\hat\prec_{c_1}\,Y,\qquad Y\,\hat\prec_{c_2}\,X.
\end{equation}
With $T_f=\infty$ the first case of~\eqref{eq:select} holds forever, and each crossing waits: $\hat h_{c_1}=X$ is not a candidate of $c_1$ before passing $c_2$, and $\hat h_{c_2}=Y$ is not a candidate of
$c_2$ before passing $c_1$. No other AGV can create an event, so none occurs.

(c) Let one AGV be on an empty floor with a directed cycle of at most $L-1$ segments. After each commitment let $\Pi$ propose once more around the cycle and then to the exit. No segment is full, so $\Psi=1$ and $R_i\le L$ throughout. Commitments never stop, so~\eqref{eq:p2} holds, and the AGV never exits, against~\eqref{eq:p3}.

(d) Let $B=1$ and let $a,b,e$ be segments, $a$ and $e$ parallel as a general floor allows, with $\mathrm{dn}(a)=\mathrm{dn}(e)=\mathrm{up}(b)$ and $\mathrm{dn}(b)=\mathrm{up}(a)=\mathrm{up}(e)$. Let $X$ on $a$ have the route $a,b,e$ and $Y$ on $e$ the route $e,b,a$. Segment $b$ is empty, so $\mathcal{W}(z)=\emptyset$. Whichever AGV enters $b$ first wishes the segment of the other, which wishes $b$, a full wait cycle. On a one-way grid the shortest cycles have four segments, and the same trap needs two of them. Let $s'$ be empty, from $c$ to $d$, and let two disjoint chains of full segments run from the two outgoing segments of $d$ back into $c$, through its two incoming segments $u$ and $v$. Every head wishes the next segment of its chain, and the heads of $u$ and $v$ wish $s'$ and then the first segment of the chain that returns through the other of $u$ and $v$. Then $\mathcal{W}(z)$ is two paths, so $z\in\mathcal{Z}_{\mathrm{safe}}$; only the heads of $u$ and $v$ can move, and whichever enters $s'$ closes one chain into a full wait cycle. On the $4\times4$ grid such chains exist, of seven and three segments.
\end{proof}

\begin{proof}[Proof of Theorem~\ref{thm:or-live}]
Suppose that $[t_0,t_0+\bar T]$ contains no exit and no commitment, and let $\mathcal{P}(t)$ be the set of commitments pending at $t$.
\begin{enumerate}
\item At $t_0$, order the pending commitments by their
original commitment times. Since disturbances have ended,
the induction in the proof of Proposition~\ref{prop:d0}
applies from the current positions and speeds: passages
already in a crossing clear within $\Delta$, and each
remaining commitment has the same per-step bound.
Thus all commitments pending at $t_0$ clear by $t_0+D_0$.
With no new commitment, every head is then ready.
\item With no commitment after $t_0$, $\tau_c(t)\ge t-t_0\ge T_f$ for every $c\in\mathcal{C}$ and $t\ge t_0+T_f$.
\item Let $t_1=t_0+\max\{T_f,D_0\}$. Then $z(t_1)\in\mathcal{Z}_{\mathrm{safe}}$ by~\eqref{eq:inv}, and Lemmas~\ref{lem:stuck} and~\ref{lem:alt}, which are statements about a single state, give~\eqref{eq:nonempty} for the ledger and routes at $t_1$, whatever routes changed before: $\mathcal{A}_c(z(t_1))\ne\emptyset$ for some $c$, its heads being ready by step~1.
\item A coordinator with period $\delta\le\min\{T_f,D_0\}$ checks at some $t'\in[t_1,t_1+\delta]$, applying the accepted route changes first and then selecting on the same ledger, to which steps~1--3 apply; by step~2, $i_c^\star\in\mathcal{A}_c$ is committed at $t'$.
\end{enumerate}
Since $t_1+\delta\le t_0+T_f+D_0=t_0+\bar T$, step~4 contradicts the assumption.
\end{proof}

\begin{proof}[Proof of Proposition~\ref{prop:d0}]

We first bound the time for one advance. Write
$a=a_{\max}$. For a head with a clear approach, take
the stop line as its target. For a follower, fix a
position $p$ already reached by the rear of its
predecessor and take $p-\gamma_0-\epsilon$ as the
target. An AGV already at or beyond its target needs
no advance. Otherwise let $D$ be the remaining
distance to the target at the start of the advance,
and let $z\in[0,D]$ be the distance travelled since.

For a follower, the available gap above $\gamma_0$ is
at least $D-z+\epsilon$ while its predecessor remains
on the segment, since the predecessor does not reverse.
If the predecessor leaves, the gap restriction disappears.
Also, the distance to the stop line is at least $D-z$.
Hence (C5) requires $\dot v=a$ whenever $v<w(z)$, where
\begin{equation}
w(z)=\min\!\left\{V,\sqrt{2a(D-z)},
                    \frac{D-z+\epsilon}{t_\gamma}\right\},
\end{equation}
with the last term omitted for a head. Since $w$ is
nonincreasing, once $v\ge w(z)$ it cannot fall below
$w(z)$ again. Before first reaching $w$, acceleration
at $a$ gives $v^2=v(0)^2+2az$. Thus throughout the
advance,
\begin{equation}
v\ge\min\{\sqrt{2az},w(z)\}.
\end{equation}

Integrating the reciprocal of this lower bound gives
the travel time. Relative to travel at speed $V$, each
of the acceleration and braking terms adds at most
\begin{equation}
\int_0^{V^2/(2a)}
\left(\frac{1}{\sqrt{2az}}-\frac1V\right)\,dz
=\frac{V}{2a}.
\end{equation}
For a follower the gap term adds at most
\begin{equation}
\int_0^D\frac{t_\gamma}{D-z+\epsilon}\,dz
=t_\gamma\ln\!\frac{D+\epsilon}{\epsilon}.
\end{equation}
Here $D+\epsilon\le\bar d$ for a follower, since
$p$ and its initial front position lie on the same
segment. A head has $D\le\bar d$. The advance therefore
takes at most $\bar d/V+V/a$ for a head and at most
$T_{\mathrm{seg}}$ for a follower.

Let $e$ be committed at $t_e$ from $s$ into $s'$ and not yet past its crossing. We show that $e$ waits only for AGVs with an earlier pending commitment.
\begin{enumerate}
\item \emph{Waits on $s$.} A crossing commits only the first uncommitted AGV of a segment, so every AGV ahead of $e$ on $s$ was committed before $t_e$.
\item \emph{Waits at the crossing.} A crossing serves its committed passages in the order of commitment, so the passages that hold $e$ by (C3) and the first part of (C4) were committed before $t_e$; the second part of (C4) holds by the placement of the stop lines and makes no AGV wait.
\item \emph{Waits for room on $s'$.} Suppose that the entry of $s'$ lacks room for $e$ once the AGVs on $s'$ have moved up. By the length condition, $B$ AGVs are then physically on $s'$, and each entered through $\mathrm{up}(s')$ ahead of $e$, hence was committed into $s'$ before $t_e$. Since $n_{s'}(t_e)<B$, one of them had been committed out of $s'$ before $t_e$; by FIFO it is the first, and $e$ waits until it passes its stop line.
\end{enumerate}
A pending AGV already inside its crossing waits for nothing by (C1) and leaves it within $\Delta$. Let $e_1,\dots,e_m=e$ be the other pending commitments at $t_e$ in the order in which they were made. Each AGV holds at most one, and these AGVs stand before their stop lines on entry or internal segments, so $m\le\bar P$. By steps~1--3, $e_k$ waits only for $e_j$ with $j<k$ and for AGVs inside crossings. We claim that $e_k$ clears its crossing by $t_e+\Delta+k(\Delta+h_0+B\,T_{\mathrm{seg}})$. For $k=1$, no earlier passage holds $e_1$ after
$t_e+\Delta$. It reaches its stop line within
$T_{\mathrm{seg}}$. To bound the time until (C1) holds,
place the first target at the downstream stop line of
the receiving segment and successive targets at
intervals $\ell+\gamma_0+\epsilon$ upstream. Once
the front of one AGV reaches its target, its rear has
reached the reference position $p$ for the next AGV.
An AGV that has left the segment has also passed its
target. Applying the single-AGV bound from front to
rear, the at most $B-1$ AGVs reach or pass their
targets within $(B-1)T_{\mathrm{seg}}$. The length
condition then gives (C1), and $e_1$ clears within
a further $h_0+\Delta$. If the claim holds for $j<k$, then $e_k$ waits for no one after $t_e+\Delta+(k-1)(\Delta+h_0+B\,T_{\mathrm{seg}})$ and clears within the same increment. Hence $e$ clears by $t_e+\Delta+\bar P(\Delta+h_0+B\,T_{\mathrm{seg}})$, and by (C5) the next AGV of $s$ becomes a candidate within $T_{\mathrm{seg}}$ more, which gives $D_0\le\Delta+T_{\mathrm{seg}}+\bar P(\Delta+h_0+B\,T_{\mathrm{seg}})\le(\bar P+1)(\Delta+h_0+B\,T_{\mathrm{seg}})$.
\end{proof}

\begin{proof}[Proof of Corollary~\ref{cor:or-clear}]
Let $\Delta\Phi$ be the change of~\eqref{eq:phi} at an event. For a commitment $\Delta R_i=-1$, and for a route change $\Delta R_i\le L-1$, since $R_i\ge1$ and $R_i^{+}\le L$, and $\Delta\kappa_i=1$, so
\begin{equation}
\Delta\Phi\big|_{\mathsf{commit}}=-1,\qquad \Delta\Phi\big|_{\mathsf{reroute}}\le(L-1)-L=-1.
\end{equation}
An exit removes a nonnegative term. With no entries, $\Phi\ge0$ and each such event lowers it by at least one, so at most $\Phi(z(T_0))$ of them occur, and Theorem~\ref{thm:or-live} keeps
them coming while $\mathcal{I}(t)\ne\emptyset$. Finally $R_i\le L$ and $\kappa_i\ge0$ give $\Phi(z(T_0))\le|\mathcal{I}(T_0)|\,L(K+1)$.
\end{proof}

\begin{proof}[Proof of Theorem~\ref{thm:advice}]
(i) Fix a crossing $c$ and let $e_1,e_2,\dots$ be the passages of the clean plan at $c$ in the order $\hat\prec_c$. We show by induction on $k$ that the $k$-th commitment at $c$ is $e_k$. Let $e_k=(i,s\to s')$, and let the first $k-1$ commitments at $c$ be $e_1,\dots,e_{k-1}$; the conditions of a clean plan hold along the run, whatever the other crossings do.
\begin{enumerate}
\item $\hat h_c=i$ until $e_k$: $e_k$ is the first passage of $\hat\prec_c$ not yet committed at $c$, and while $i$ is a head its move has $\mathrm{Look}=1$ by the definition of a clean plan, so $i$ is never a refused head.
\item The first case of~\eqref{eq:select} applies whenever $c$ decides before $e_k$: the commitment before $e_k$ at $c$ is served and $i$ is admitted less than $T_f$ after it, or after $t=0$ with $\tau_c(0)=0$, so $\tau_c<T_f$.
\item By steps 1 and 2, $c$ commits nothing while $i\notin\mathcal{A}_c$ and selects $i$ once $i\in\mathcal{A}_c$; no later AGV of $\hat\prec_c$ is committed before $i$.
\item The route changes of $i$ are accepted, since $\kappa_i<K$ and $\Psi=1$ in the same state.
\end{enumerate}
Hence the $k$-th commitment at $c$ is $e_k$, which completes the induction.

(ii) Let $C(T)$ and $E(T)$ count the commitments and exits by $T$. By Theorem~\ref{thm:or-live},
\begin{equation}
C(T)+E(T)\ge\lfloor T/\bar T\rfloor.
\end{equation}
An AGV starts with $R_i\le L-1$, so the term of $i$ in~\eqref{eq:phi} starts at most at $L(K+1)-1$. Each commitment lowers it by one, so $i$ makes at most $L(K+1)$ commitments and exits together. Counting an exit at the commitment into the exit segment, as in~\eqref{eq:event}, advances it by a bounded time and leaves the limit~\eqref{eq:thr} unchanged. Every AGV counted in $C$ or $E$ has exited or is among the at most $N=B|\mathcal{S}_{\mathrm{int}}|$ AGVs on the internal segments, so
\begin{equation}
C(T)+E(T)\le L(K+1)\bigl(E(T)+N\bigr).
\end{equation}
Dividing by $T$ and letting $T\to\infty$ gives~\eqref{eq:robust}.
\end{proof}

\begin{proof}[Proof of Proposition~\ref{prop:np}]
The walk~\eqref{eq:walk} visits each segment at most once. For the second claim we reduce from the job shop $J3\,|\,p_{ij}=1\,|\,C_{\max}$, which is NP-hard~\cite{lenstra1979computational}. Let the machines be three crossings, and let each job be an AGV. Give each job a private entry, a private segment between each two consecutive operations, and a private exit, so that its route visits the crossings in the order of its operations. Each segment then carries one AGV, so $n_s\le1<B$, $\mathcal{F}=\emptyset$, and $\mathrm{Look}\equiv1$. A crossing serves one AGV per unit time, no two share a segment, and traversal times are negligible. So schedules of the job shop and commitment sequences correspond one to one with the same makespan.
\end{proof}

\subsection{Price of a guarantee}\label{app:price}

\begin{proof}[Proof of Theorem~\ref{thm:flow}]
Suppose $N_s(t)>B$ and let $d\in\mathcal{D}^\circ$ stop the AGVs inside $s$ from $t$ on. The AGVs past the first stop line cannot turn back, and $s$ holds only $B$, so an AGV stops in the
crossing, against robustness. Hence $N_s(t)\le B$. Counting passages against the integral of $N_s$ then gives~\eqref{eq:int2}, which with~\eqref{eq:ts} is~\eqref{eq:flow}.
\end{proof}

\begin{proof}[Proof of Proposition~\ref{prop:tight}]
On a lane no chain closes, so $\chi=0$ and every move into a segment with room is admitted. Work conservation makes~\eqref{eq:maxplus} tight in its largest term,
\begin{equation}
g_j(k)=\max\{g_{j-1}(k)+B+1,\,g_{j+1}(k-B)+\omega,\,g_j(k-1)+1\},
\end{equation}
a max-plus linear system. Its circuits are $j\to j+1\to j$ through~\eqref{eq:mp1},\eqref{eq:mp2}, of weight $B+1+\omega$ over $B$ AGVs, and the loop~\eqref{eq:mp3}, of mean one. The system is
irreducible for at least two segments, so by the cycle-time theorem
\begin{equation}
\lim_{k\to\infty}\frac{g_j(k)}{k}=\max\Bigl\{\frac{B+1+\omega}{B},\ 1\Bigr\}=\frac{B+1+\omega}{B}
\end{equation}
from any initial state. The flow $\lambda$ is the inverse, which is~\eqref{eq:tight}.
\end{proof}

\begin{proof}[Proof of Proposition~\ref{prop:ring}]
The recursion on the ring is max-plus linear with one node per segment, and irreducible for $Q<MB$, so the cycle time is the largest mean over simple circuits. A simple circuit of a ring with arcs in both directions is one of four kinds: a loop~\eqref{eq:mp3}, of mean one; a pair $j\to j+1\to j$, of mean $(B+1+\omega)/B$; a forward lap, of mean $M(B+1)/Q$ by~\eqref{eq:ringv}; and a backward lap, of mean $M\omega/(MB-Q)$ by~\eqref{eq:ringh}. The pair mean is the mediant of the two lap means,
\begin{equation}
\frac{B+1+\omega}{B}=\frac{M(B+1)+M\omega}{Q+(MB-Q)},
\end{equation}
so it lies between them and never exceeds the larger, and the loop mean is below it. The inverse of the larger lap mean is~\eqref{eq:ringfd}.
\end{proof}

\section{Expanded Derivations}\label{app:deriv}

\subsection{Flow bound as an occupancy integral}

Start from an empty floor. Let $[a_p,b_p)$ be the $p$-th passage of $s$, from its first stop line to its second. Since the speed is at most $V$,~\eqref{eq:ts} gives
\begin{equation}\label{eq:pass}
b_p-a_p\ge T_s,
\end{equation}
and by the definition of $N_s$,
\begin{equation}\label{eq:indicator}
N_s(t)=\sum_{p}\mathbf{1}\{t\in[a_p,b_p)\}.
\end{equation}
Let $D(\tau)=|\{p:b_p\le\tau\}|$ count the passages completed by $\tau$. Each has $[a_p,b_p)\subseteq[0,\tau]$. Integrating~\eqref{eq:indicator}, dropping the passages still open at $\tau$, and using~\eqref{eq:pass},
\begin{align}
\int_0^{\tau}N_s(t)\,dt
&=\sum_{p}\bigl|[a_p,b_p)\cap[0,\tau]\bigr| \nonumber\\
&\ge\sum_{p:\,b_p\le\tau}(b_p-a_p)\ \ge\ D(\tau)\,T_s. \label{eq:int1}
\end{align}
With $N_s\le B$ the left side is at most $B\tau$, so
\begin{equation}\label{eq:int2}
\frac{D(\tau)}{\tau}\ \le\ \frac{1}{\tau T_s}\int_0^{\tau}N_s(t)\,dt\ \le\ \frac{B}{T_s},\qquad\forall\,\tau>0.
\end{equation}

\subsection{Circuits of the ring}\label{app:circuit}

Let an event graph be a set of inequalities $g_{v}(k)\ge g_{u}(k-m_{uv})+w_{uv}$, with $m_{uv}\in\mathbb{Z}_{\ge0}$ tokens and weight $w_{uv}\ge0$. Along a circuit $\mathcal{O}=(v_0,\dots,v_p=v_0)$ put
\begin{equation}\label{eq:mw}
m_{\mathcal{O}}=\sum_{l=0}^{p-1}m_{v_lv_{l+1}},\qquad W=\sum_{l=0}^{p-1}w_{v_lv_{l+1}}.
\end{equation}
Chaining the $p$ inequalities,
\begin{equation}\label{eq:chain2}
g_{v_0}(k)\ge g_{v_0}(k-m_{\mathcal{O}})+W.
\end{equation}
If $m_{\mathcal{O}}\ge1$, iterating~\eqref{eq:chain2} and dividing by $k$ gives $\liminf_k g_{v_0}(k)/k\ge W/m_{\mathcal{O}}$, so
\begin{equation}\label{eq:circuit}
\lambda\le\frac{m_{\mathcal{O}}}{W}.
\end{equation}
 If $m_{\mathcal{O}}=0$ and $W>0$, then~\eqref{eq:chain2} reads $g_{v_0}(k)\ge g_{v_0}(k)+W$, which no finite time satisfies: the event never occurs. We substitute the ring, whose circuits are those used in the proof of Proposition~\ref{prop:ring}.

\emph{A ring.} Let $Q$ AGVs circulate on a ring of $M$ segments, and index segments by $j\in\mathbb{Z}$ along the direction of travel. After one lap an AGV returns to the same segment $Q$ entries later, so
\begin{equation}\label{eq:wrap}
g_{j+M}(k)=g_j(k+Q).
\end{equation}
Applying~\eqref{eq:mp1} $M$ times and substituting~\eqref{eq:wrap} gives the circuit of the vehicles,
\begin{equation}\label{eq:ringv}
g_j(k+Q)\ge g_j(k)+M(B+1),
\end{equation}
with $Q$ tokens. Applying~\eqref{eq:mp2} $M$ times gives $g_j(k)\ge g_{j+M}(k-MB)+M\omega$, and~\eqref{eq:wrap} turns this into the circuit of the holes,
\begin{equation}\label{eq:ringh}
g_j(k)\ge g_j(k-(MB-Q))+M\omega,
\end{equation}
with $MB-Q$ tokens, the free places of the ring. At $Q=MB$ circuit~\eqref{eq:ringh} has no token, and for $\omega>0$ no AGV moves: this is the full wait cycle of Definition~\ref{def:fwc}, and one hole is the least that gives the circuit a token.

\bibliographystyle{IEEEtran_doi}
\bibliography{refs}

\ifblind\else
\vspace{-15mm}
\begin{IEEEbiography}[{\includegraphics[width=1in,height=1.25in,clip,keepaspectratio]{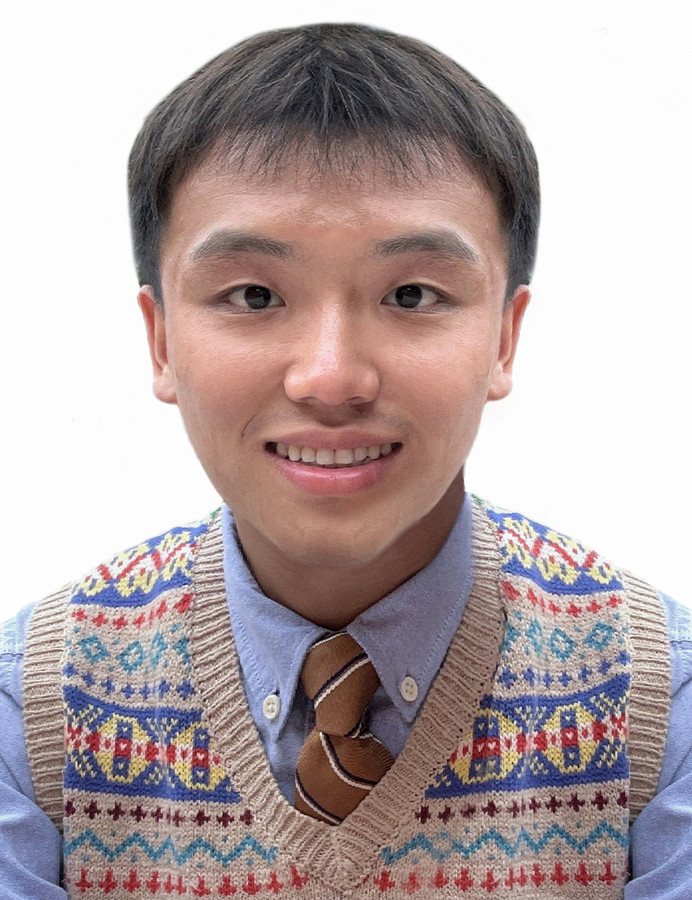}}]{Zhihao Lin}
received the Ph.D. degree in autonomous systems and connectivity from the University of Glasgow, Glasgow, U.K., in
2026. He is currently a JSPS Postdoctoral Fellow at the Graduate School of Information Science and Technology, The
University of Osaka, Japan. His research interests include multi-agent coordination, game theory, and safe autonomous driving.
\end{IEEEbiography}

\vspace{-20mm}
\begin{IEEEbiography}[{\includegraphics[width=1in,height=1.25in,clip,keepaspectratio]{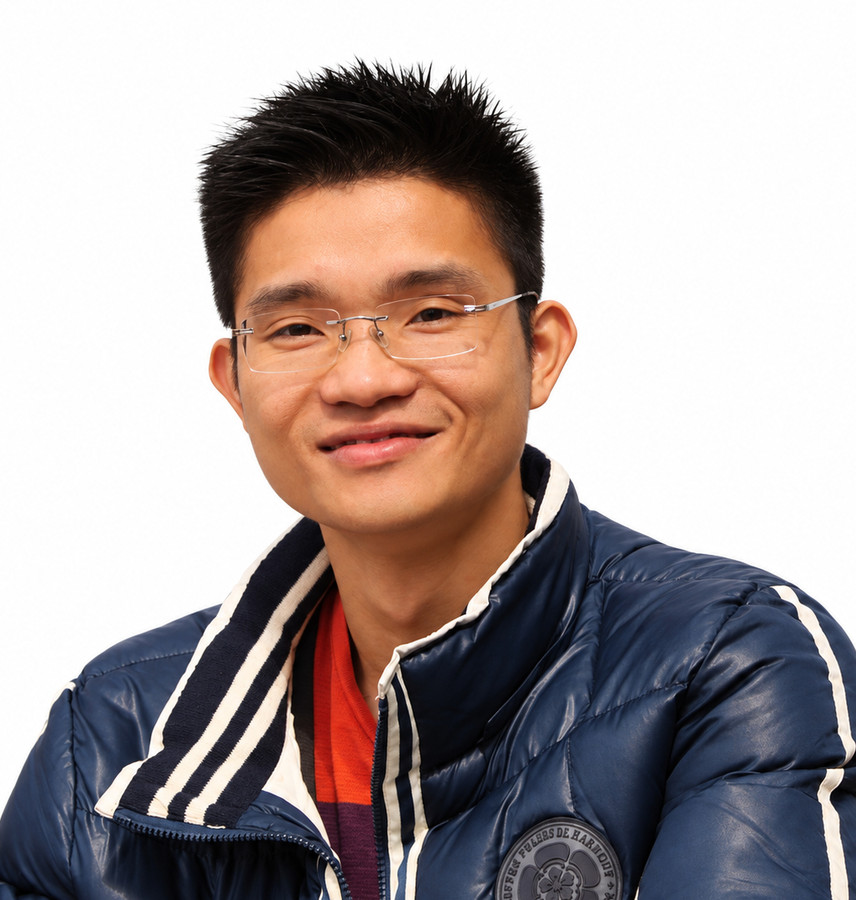}}]{Anh-Tu Nguyen}
(Senior Member, IEEE) received the Ph.D. degree in automatic control from the Universit\'e Polytechnique
Hauts-de-France, Valenciennes, France, in 2013. He is currently an Associate Professor with INSA Hauts-de-France,
Universit\'e Polytechnique Hauts-de-France. His research interests include robust control and estimation, cybernetic
control systems, and mechatronic applications. He serves as an Associate Editor for the \textsc{IEEE/ASME
Transactions on Mechatronics}, the \textsc{IEEE Transactions on Automation Science and Engineering}, the
\textsc{IEEE Transactions on Intelligent Transportation Systems}, \emph{Control Engineering Practice},
\emph{Engineering Applications of Artificial Intelligence}, and \emph{ISA Transactions}.
\end{IEEEbiography}

\vskip 0pt plus -1fil
\begin{IEEEbiography}[{\includegraphics[width=1in,height=1.25in,clip,keepaspectratio]{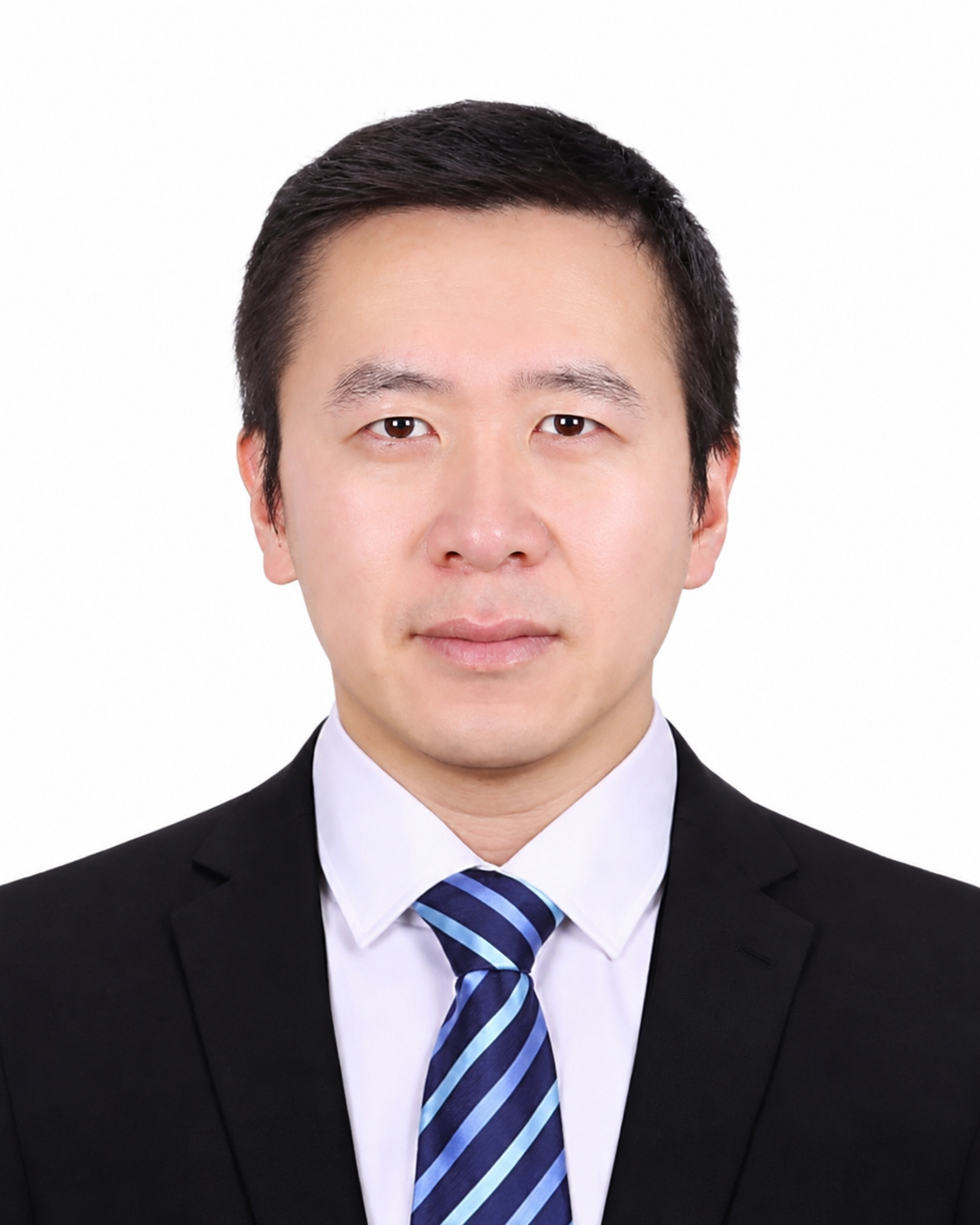}}]{Hui Zhang}
(Fellow, IEEE) received the B.Sc. degree in mechanical design manufacturing and automation from the Harbin Institute of Technology, Weihai, China, in 2006, the M.Sc. degree in automotive engineering from Jilin University, Changchun, China, in 2008, and the Ph.D. degree in mechanical engineering from the University of Victoria, Victoria, BC, Canada, in 2012. He has been a Professor with the School of Transportation Science and Technology, Beihang University, Beijing, China, since 2016. His research interests include special vehicle design, vehicle dynamics control, and intelligent driving.

Dr. Zhang is a member of SAE International and ASME. He received the 2017 \textsc{IEEE Transactions on Fuzzy Systems} Outstanding Paper Award, the 2018 SAE Ralph R. Teetor Educational Award, the IEEE Vehicular Technology Society 2019 Best Vehicular Electronics Paper Award, the 2022 IEEE IES Early Career Award, and the 2025 SAE Forest R. McFarland Award.
\end{IEEEbiography}

\vskip 0pt plus -1fil
\begin{IEEEbiography}[{\includegraphics[width=1in,height=1.25in,clip,keepaspectratio]{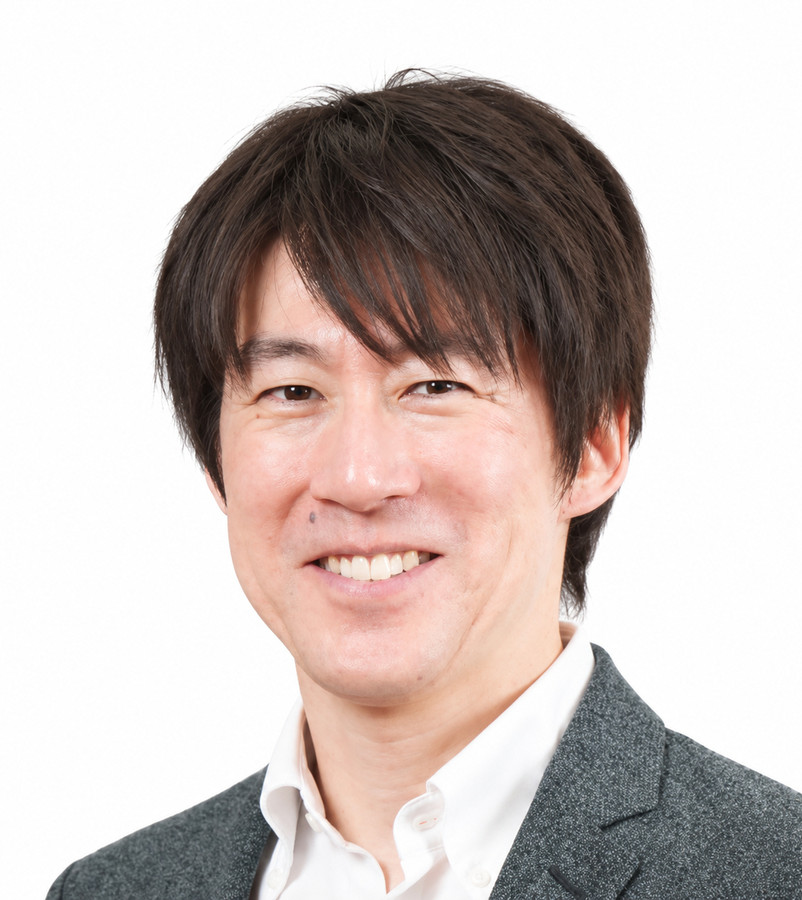}}]{Yoshinobu Kawahara}
received the B.S., M.S., and Ph.D. degrees in engineering from the University of Tokyo, Japan, in 2003, 2005, and
2008. He is a Professor with the Graduate School of Information Science and Technology, The University of Osaka,
Japan, Team Director of the Dynamical Systems Learning Team at the RIKEN Center for Advanced Intelligence Project,
and a Project Manager of the JST Moonshot R\&D Program (Goal 3). He was previously a Professor with the Institute of
Mathematics for Industry, Kyushu University. He received the Young Scientists' Prize of the MEXT Commendation for
Science and Technology in 2020, and serves as an Action Editor for \emph{Transactions on Machine Learning Research}
and \emph{Neural Networks}. His research interests lie at the intersection of machine learning and dynamical
systems, including operator-theoretic methods and their applications.
\end{IEEEbiography}
\fi

\end{document}